\documentclass[a4paper, twoside]{article}

\usepackage{epsfig}
\usepackage{subfigure}
\usepackage{calc}
\usepackage{amssymb}
\usepackage{amstext}
\usepackage{amsmath}
\usepackage{amsthm}
\usepackage{multicol}
\usepackage{pslatex}
\usepackage{apalike}
\usepackage{SciTePress}
\usepackage[small]{caption}
\usepackage{enumerate} 
\usepackage{xfrac}
\usepackage{scrextend}
\usepackage{verbatim}
\usepackage{etoolbox}
\makeatletter
\patchcmd{\verbatim@input}{\@verbatim}{\scriptsize\@verbatim}{}{}
\makeatother
\usepackage{graphicx}

\newtheorem{theorem}{Theorem}
\newtheorem{corollary}{Corollary}
\newtheorem{proposition}{Proposition}

\begin{document}

\title{\uppercase{ An Elliptic Curve Based Solution to the Perspective-Three-Point Problem }}

\author{\authorname{Michael Q. Rieck\sup{1}}
\affiliation{\sup{1}Mathematics and Computer Science Department, Drake University, Des Moines, IA 50310, USA}
\email{mrieck@drake.edu}
}


\abstract{The Perspective-Three-Point Problem (P3P) is solved by first focusing on determining the directions of the lines through pairs of control points, relative to the camera, rather than the distances from the camera to the control points. The analysis of this produces an efficient, accurate and reasonably simple P3P solver, which is compared with a state-of-the-art P3P solver, ``Lambda Twist." Both methods depend on the accurate computation of a single root of a cubic polynomial. They have been implemented and tested for a wide range of control-point triangles, and under certain reasonable restrictions, the new method is noticably more accurate than Lambda Twist, though it is slower. However, the principal value of the present work is not in introducing yet another P3P solver, but lies rather in the discovery of an intimate connection between the P3P problem and a special family of elliptic curves that includes curves utilized in cryptography. This holds the potential for further advances in a number of directions. To make this connection, an interesting spherical analogue of an ancient ``sliding" problem is stated and solved. 
}

\onecolumn \maketitle \normalsize \vfill

\section{\uppercase{Introduction}}
\label{sec:Introduction}

While the Perspecive-Three-Point (Pose) Problem has been of mathematical and practical interest for most of two centuries, new insights into its nature continue to be discovered, and increasingly better techniques for solving it continue to be developed. The problem, as introduced and solved in \cite{G}, involves using the images in a photograph of three known points in space (the ``control points") to determine the pose (position and orientation) of the camera that took the photograph. As is rather well known, this problem typically does not lead to a unique mathematical solution, but rather, may yield up to four distinct solutions, any of which could have been the pose of the camera. The P3P problem has subsequently been applied to the area of camera tracking and is considered one of its foundational problems.

By the end of the twentieth century, quite a few ``P3P solvers," had been developed. In the comprehensive study \cite{HLON}, these various algorithms for solving the P3P problem were systematically classified and compared.  In \cite{GHT}, the nature of various possible solutions to the P3P problem, based on the parameters of the problem, were carefully characterized. Another P3P solver was introduced as well. Currently, among the most successful (most accurate, most reliable, most efficient) P3P solvers are discussed in \cite{KR}, \cite{KSS}, \cite{WXWC} and \cite{PN}.  

The present research article introduces yet another P3P solver and reports on tests made comparing it to the ``Lambda Twist" method of \cite{PN}. However, the principal focus of this article is not on promoting yet another P3P solver, but is rather on making solid and interesting connections between three seemingly very different mathematical topics: 

\begin{enumerate}

\item a problem concerning the sliding of an arc of a great circle around a sphere, 

\item the P3P problem, and 

\item elliptic curves and related genus-one curves. 

\end{enumerate} 

Section 2 introduces and solves the arc-sliding problem. The solution to this is a quartic curve on the sphere that can be easily converted to a quartic curve in the real projecting plane. Section 3 investigates the nature of the singularities of this curve. Because of the ease in visualizing the arc-sliding problem, these singularities are quickly perceived to be double points on the curves. The curve in the projective plane can then be ``adjusted" via a projective transformation to put it into a somewhat more pleasing and ``standard" form.  

Section 4 details precisely how the P3P problem is related to the arc-sliding problem. It is in fact equivalent to finding the points of intersection of either of the above two quartic curves with a corresponding straight line. This provides a potentially useful way to interpret the P3P problem in terms of such curves in the projective plane. However, reinterpreting P3P in terms of projective plane curves is not a new idea. In fact, the method of Finsterwalder,  the method of Grafarend et al., both discussed in \cite{HLON}, and more recently, the Lambda Twist method, essentially begin by reducing the P3P problem to the problem of finding the intersection of two conic sections, in the projective plane. 

Section 5 introduces a substantial family of quartic polynomials in two variables and their associated (genus-one) curves. This family of curves is broad enough to include the curves obtained in Section 3, as well as the so-called ``Edwards curves" and ``twisted Edwards curves" that have become useful in the area of elliptic-curve cryptography. This section includes a birational transformation that can be applied to convert a typical member of the family of curves to a classical Jacobi quartic (elliptic) curve. 

Section 6 describes the details of a new algorithm for solving the P3P problem based on the ideas of the preceding sections. It also explains many of the details of a C computer program that implements this algorithm. Section 7 goes on to describe a number of tests of this C code in contrast with some C code that implements the Lambda Twist P3P solver and which is mostly just a translation of previously existing C++ code. Several tests were conducted for the purpose of comparing the two algorithms. The details of the test designs and the results of the tests are laid out in Section 7. While the new algorithm is considerably slower than Lambda Twist, it is nevertheless more accurate under reasonable assumptions. Section 8 sums up the results of the investigation into applying elliptic curves to the P3P problem. 

\section{\uppercase{An Arc-Sliding Problem}}
\label{sec:An Arc-Sliding Problem}

The setup to be considered in this section generalizes some ancient mathematical ``sliding" puzzles as discussed in the following cited articles. Theorem 1 of \cite{W} states that if a triangle moves around in a plane so as to keep two of its vertices on prescribed intersecting lines, then the third vertex will trace out an ellipse. If we replace the triangle  with a line segment, constrained to move with its endpoints on prescribed intersecting lines, and if we focus on the motion of some point on this line segment, then the curve traced out by this point will again be an ellipse. As discussed in \cite{AM}, when the prescribed lines are perpendicular, this fact serves as a mathematical basis for an ancient and still popular ellipse-drawing tool known as the Trammel of Archimedes.  

The generalization to be presented here moves the venue for such sliding problems from the flat plane to the unit sphere. In the limit as the line segments involved become small, the spherical analogues reduce to the original flat-plane problems. In the spherical analogues, straight lines and line segments are replaced with great circles and arcs along great circles. 

Our specific interest will be on sliding an arc around the sphere so as to keep its endpoints constrained to lie on prescribed great circles. As the arc moves around in this way, the first problem to be addressed is that of finding a description of the curve on the sphere traced out by the motion of some given point on the arc. 

Technically, the analysis to be provided would also allow this point to lie somewhere outside the arc but still on the (moving) great circle containing the arc. However, to make the discussion smoother, it will be assumed, at least in this section, that the point is on the moving arc, not beyond it. For the same reason, it will be assumed that the arc has length less than $\pi$, and so can be contained in a hemisphere. 

As we will see, generally speaking, the curve constructed in this way on the unit sphere is given by a fourth degree polynomial in the Cartesian coordinates. This spherical curve leads naturally to a rather interesting fourth-degree (quartic) curve in the real projective plane $\mathbb{RP}_2$, as follows. Through each point of the curve on the sphere, consider the line in real three-dimensional space $\mathbb{R}^3$ that passes through this point and the origin. By definition, this line {\it is} a point of $\mathbb{RP}_2$ and the totality of such points form a curve in $\mathbb{RP}_2$.

\begin{figure} 
\centerline{ \includegraphics[width=7cm]{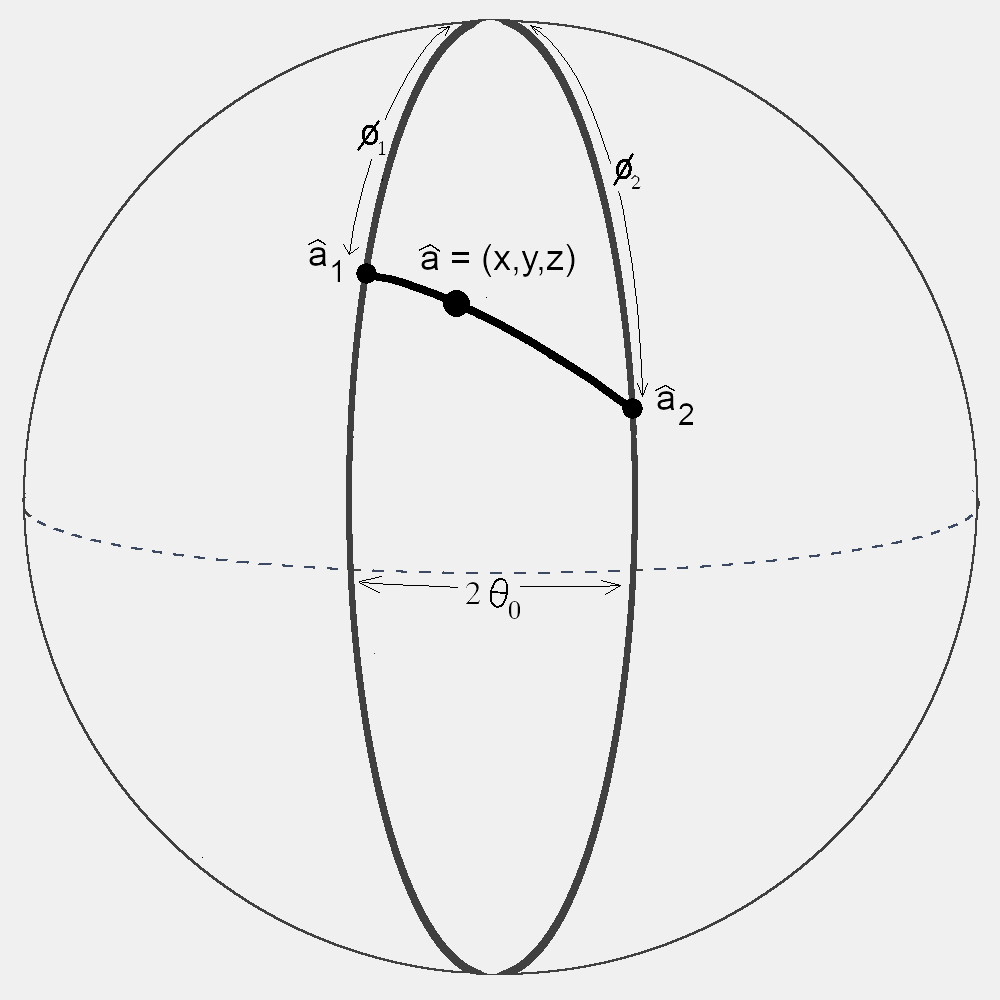}  }
\caption{an arc-sliding problem  } 
\label{fig:1}
\end{figure}

Besides using Cartesian coordinates in $\mathbb{R}^3$, it will help to consider spherical coordinates on the unit sphere. The azimuthal angle (``longitude") will be denoted by $\theta$. The polar angle (``latitude" as measured from the ``north pole") will be denoted $\phi$. Thus, on the unit circle, $x = \cos \theta \sin \phi$, $y = \sin \theta \sin \phi$ and $z = \cos\phi$.  Owing to the symmetry of the unit sphere, there is no loss of generality in assuming that the two fixed great circles in the problem are the $\theta = - \theta_0$ and the $\theta = \theta_0$ circles, for some fixed number $\theta_0$ with $0 < \theta_0 < \pi / 2$. These are the intersections of the unit sphere with the planes $\sin \theta_0 \; x + \cos \theta_0 \; y = 0$ and $\sin \theta_0 \; x - \cos \theta_0 \; y = 0$, respectively. It will be convenient to introduce the following notation: 

\vspace{-2mm}

$$\mu_0 = \cos \theta_0 \; \; \hbox{and} \; \; \nu_0 = \sin \theta_0. $$

\vspace{2mm}

The two points that slide along the great circles will be represented by two unit vectors $\hat{a}_1$ and $\hat{a}_2$. For convenience, no distinction will be made between points on the unit sphere and the vectors that represent them. The Cartesian coordinates for $\hat{a}_1$ and $\hat{a}_2$ are as follows: 

% \vspace{-2mm}

$$\hat{a}_1 \; = \; ( \, \mu_0 \, \nu_1, \; -\nu_0 \, \nu_1, \; \mu_1 \, ) \; \; \; \hbox{and}$$ 

\vspace{-4mm}

$$\hat{a}_2 \; = \; ( \, \mu_0 \, \nu_2, \; \nu_0 \, \nu_2, \; \mu_2 \, ), \; \; \hbox{where}$$

\vspace{-6mm}

$$\mu_1 = \cos \phi_1, \; \nu_1 = \sin \phi_1, \; \mu_2 = \cos \phi_2, \; \nu_2 = \sin \phi_2, $$

\vspace{2mm}

\noindent and $\phi_i$  is a varying number with $0 \le \phi_i \le \pi$ \, ($i = 1,2$). The dot product $\hat{a}_1 \cdot \hat{a}_2$ will be assumed to stay constant as $\hat{a}_1$ and $\hat{a}_2$ slide along their respective great circles. Notice that this constant is the cosine of the length (assumed less than $\pi$) of the arc connecting $\hat{a}_1$ and $\hat{a}_2$ along a great circle. Notice too that 

\vspace{3mm}

\noindent $$\hat{a}_1 \cdot \hat{a}_2 \; \; = \; \; (\mu_0^2 - \nu_0^2) \, \nu_1 \nu_2 \; + \; \mu_1 \mu_2.$$

\vspace{2mm}

The unit vector $\hat{a}$ will represent the point whose path we are interested in tracing as $\hat{a}_1$ and $\hat{a}_2$ slide up and down the great circles. It is assumed that $\hat{a}$ lies on the (moving) great circle containing both $\hat{a}_1$ and $\hat{a}_2$. Therefore, 

\vspace{-2mm}

$$ \hat{a} \; \; = \; \; \alpha_1 \; \hat{a}_1\; +  \; \alpha_2 \; \hat{a}_2, $$

\vspace{1mm}

\noindent where $\alpha_1$ and $\alpha_2$ are fixed real numbers. To ensure that $\hat{a}$ is a unit vector, it is also required that 

\vspace{-2mm}

$$ \alpha_1^2 \; + \; \alpha_2^2 \; + \; 2 \, \alpha_1 \alpha_2 \; \hat{a}_1 \cdot \hat{a}_2  \; \; = \; \; 1.$$

\vspace{2mm}

\noindent Assuming that $\alpha_1 \ge 0$ and  $\alpha_2 \ge 0$, it is necessary that $\alpha_1 + \alpha_2 \ge 1$ and $|\alpha_1 - \alpha_2| \le 1$. (Similar restrictions are easily derived  when the alphas are allowed to be negative.)  

Since our primary focus is on the movement of $\hat{a}$, we henceforth limit the usage of the symbols $x$, $y$ and $z$ to the Cartesian coordinates of this point. The entire setup, which can be seen in Figure 1, can now basically be described via the following system of equations: 

\vspace{-4mm}

\begin{equation}
\end{equation}
$$
\left\{ \begin{array}{lll}
x \; & \;  = \; & \; \alpha_1 \, \mu_0 \, \nu_1 \; + \;  \alpha_2 \, \mu_0 \, \nu_2 \\
y \; & \; = \; & \; -\alpha_1 \, \nu_0 \, \nu_1 \; + \;  \alpha_2 \, \nu_0 \, \nu_2 \\
z \; & \; = \; & \; \alpha_1 \, \mu_1 \; + \;  \alpha_2 \, \mu_2 \\
1 \; & \; = \; & \; \alpha_1^2 \; + \alpha_2^2 \; + \\
\ & \ & \; \; 2 \, \alpha_1 \alpha_2 \; [ \, (\mu_0^2 - \nu_0^2) \, \nu_1 \nu_2 \; + \; \mu_1 \mu_2 \, ]. \\
\end{array} \right.
$$

\vspace{2mm}

The plan now is to eliminate $\mu_1$, $\nu_1$, $\mu_2$ and  $\nu_2$ to obtain a polynomial equation in $x$, $y$ and $z$ whose coefficients involve the constant parameters $\mu_0$, $\nu_0$, $\alpha_1$ and $\alpha_2$.  Since the first two equations are linear in $\nu_1$ and $\nu_2$, it is straightforward to obtain 

\vspace{-2mm}

\begin{equation}
\nu_1 \; \; = \; \; \frac{\nu_0 \, x - \mu_0 \, y}{2 \alpha_1 \mu_0 \nu_0} \; \; \hbox{and} \; \; \nu_2 \; \; = \; \; \frac{\nu_0 \, x + \mu_0 \, y}{2 \alpha_2 \mu_0 \nu_0} \, {\ \atop .}
\end{equation}

\vspace{2mm}

\noindent When these formulas are substituted into the last equation in the system (1), the result can be written as a rational expression whose numerator must be zero. This yields the following: 

\vspace{-2mm}

\begin{equation} 
\begin{array}{l}
2 \, \mu_0^2 \nu_0^2 \, (1-\alpha_1^2-\alpha_2^2 - 2 \alpha_1 \alpha_2 \; \mu_1 \mu_2) \; \; = \\ \ \\ 
(\mu_0^2-\nu_0^2)(\nu_0^2 \, x^2 - \mu_0^2 \, y^2). 
\end{array}
\end{equation} 

\vspace{2mm}

\noindent This can be solved for $\mu_1 \mu_2$ and squared to get a formula for  $\mu_1^2 \, \mu_2^2$. Using (2), we also obtain a formula for $(1-\nu_1^2)(1-\nu_2^2)$, But of course, $\mu_1^2 \, \mu_2^2 = (1-\nu_1^2)(1-\nu_2^2)$. Out of all this, the sought-after polynomial equation can be produced, which is as follows. \\ 

\begin{theorem} 

Assuming the preceding setup, the point $\hat{a}$ with Cartesian coordinates $(x, y, z)$ moves along a curve on the unit sphere given by the equation $x^2 + y^2 + z^2 \; = \; 1$ together with the equation    

\vspace{-2mm}

\begin{equation}
\end{equation}
$$ \begin{array}{ll}
\ & [ \, 1 - (\mu_0^2 - \nu_0^2)^2 \, ] \, [ \, \nu_0^2 \; x^2 - \mu_0^2 \; y^2 \, ]^2 \\ \ \\ 
+ \; & \; 4 \,  (1-\alpha_1^2-\alpha_2^2) \, \mu_0^2 \, \nu_0^2 \, (\mu_0^2 - \nu_0^2) (\nu_0^2 \; x^2 \, - \, \mu_0^2 \; y^2) \\ \ \\ 
- \; & \; 4 \, ( \, \alpha_1^2 + \alpha_2^2 \, )  \, \mu_0^2 \, \nu_0^2 \, (\nu_0^2 \; x^2 \, + \, \mu_0^2 \; y^2) \\ \ \\ 
+ \; & \; 8 \, (\alpha_2^2 - \alpha_1^2) \, \mu_0^3 \, \nu_0^3 \; x \; y \\ \ \\ 
- \; & \; 4 [ 1 - (\alpha_1+\alpha_2)^2 ]  [ 1 - (\alpha_1-\alpha_2)^2 ] \, \mu_0^4 \, \nu_0^4 \; \; = \; \; 0.
\end{array} $$

\end{theorem} 

\vspace{4mm}

\noindent Although not needed until Section 6, it is worth noting here that $\mu_1$ and $\mu_2$ can be expressed as rational functions of $x$, $y$ and $z$, as follows: 

\vspace{-2mm}

\begin{equation}
\mu_1 \; = \; \frac{\tau_+}{4 \, \alpha_1 \, \mu_0^2 \, \nu_0^2 \; z} \; \; \hbox{and} \; \; \mu_2 \; = \; \frac{\tau_-}{4 \, \alpha_2 \, \mu_0^2 \, \nu_0^2 \; z} \, ,
\end{equation} 

%\vspace{2mm}

\noindent where $\tau_\pm \; =$

\vspace{-2mm}

$$ \begin{array}{l} 
(\nu_0^2-\mu_0^2-1) \nu_0^2 \; x^2 + (\mu_0^2-\nu_0^2-1) \mu_0^2 \; y^2 \pm 2\mu_0 \nu_0 \; x \, y \\ \ \\
+ \; 2\, (1 \pm \alpha_1^2 \mp \alpha_2^2) \; \mu_0^2 \, \nu_0^2. 
\end{array} $$

\vspace{2mm}

\noindent To obtain the formula for $\mu_1$, solve the third equation in (1) for $\mu_2$, and combine this with the formula for $\mu_1  \mu_2$ obtained from (3). Then substitute $1 - \nu_1^2$ for $\mu_1^2$, and then use (2). This produces a linear equation in $\mu_1$, that does not involve $\mu_2$, $\nu_1$ nor $\nu_2$.  Similarly, the formula for $\mu_2$ can be derived.  Another result that is needed later is as follows. 

\vspace{2mm}

\begin{proposition} 

$$\alpha_1 \; = \; \frac{ \hat{a}_1 \cdot \hat{a}  \; - \; (\hat{a}_1 \cdot \hat{a}_2)(\hat{a}_2 \cdot \hat{a})  }{ 1 \; - \; (\hat{a}_1 \cdot \hat{a}_2)^2  } $$

\noindent and 

$$\alpha_2 \; = \; \frac{ \hat{a}_2 \cdot \hat{a}  \; - \; (\hat{a}_1 \cdot \hat{a}_2)(\hat{a}_1 \cdot \hat{a})  }{ 1 \; - \; (\hat{a}_1 \cdot \hat{a}_2)^2  } $$

\end{proposition} 

\begin{proof} 

Beginning with $\hat{a}_1 \cdot \hat{a}$ \ = \ $\hat{a}_1 \cdot ( \alpha_1 \; \hat{a}_1\; +  \; \alpha_2 \; \hat{a}_2)$ \ = \ $\alpha_1 \; + \; \alpha_2 \, (\hat{a}_1 \cdot \hat{a}_2)$, and similarly,  $\hat{a}_2 \, \cdot \hat{a}$ \ = \  $(\hat{a}_1 \cdot \hat{a}_2) \, \alpha_1 \; + \; \alpha_2$, the result follows via some linear algebra. 

\end{proof}

We will now obtain the equation for the curve in the projective plane that corresponds to the curve (4) on the sphere.  This is {\it not} merely a matter of homogenizing (4) in the usual way. Since $\hat{a}$ is a point on the spherical curve, with coordinates $(x,y,z)$, letting $(X, Y, Z)$ be the coordinates of a generic point on the line through the origin and $\hat{a}$, these coordinates are proportional, {\it i.e.} $X : Y : Z \; = \, x : y : z$.  So $x \; = \, x / \sqrt{x^2+y^2+z^2} \; = \; X / \sqrt{X^2+Y^2+Z^2}$, and similarly,  $y \; = \; Y / \sqrt{X^2+Y^2+Z^2}$. The equation for the curve in the projective plane is obtained from (4) by substituting $X / \sqrt{X^2+Y^2+Z^2}$ and $Y / \sqrt{X^2+Y^2+Z^2}$ for $x$ and $y$, respectively, simplifying this to get a rational expression equal to zero, and then setting its numerator to zero. This is a fairly laborious process (without the aid of algebra manipulation software), but the resulting homogeneous equation is not too complicated. \\

\begin{corollary}

The equation of the curve in the projective plane corresponding to the curve on the unit sphere given by (4) is as follows: 

\vspace{-4mm}
\begin{equation}
\end{equation}
\vspace{-6mm}
$$ \begin{array}{lll}
0 & \, = \, & (\alpha_1^2 - \alpha_2^2)^2 \, \mu_0^2 \, \nu_0^2 \; (X^4 + Y^4) \\ \ \\
\ & + \, & \, [ \; 1 + 2 \, (\alpha_1^2 - \alpha_2^2)^2 \, \mu_0^2 \, \nu_0^2 \; ] \; X^2 \;Y^2  \\ \ \\
\ & + \, & \, [ \; 1 - 2 \, \beta \, \mu_0^2  \; ] \, \nu_0^2 \; X^2 \, Z^2 \\ \ \\
\ & + \, & \, [ \; 1 - 2 \, \beta \, \nu_0^2   \; ] \, \mu_0^2 \;Y^2 \, Z^2 \\ \ \\
\ & + \, & \, 2 \, (\alpha_1^2 - \alpha_2^2) \, \mu_0 \, \nu_0 \; X \, Y \, (X^2 + Y^2 + Z^2) \\ \ \\ 
\ & + \, & \,  [ 1 - (\alpha_1+\alpha_2)^2 ]  [ 1 - (\alpha_1-\alpha_2)^2 ] \, \mu_0^2 \, \nu_0^2 \; Z^4.  
\end{array} $$

\vspace{2mm}

\noindent where $\beta = \alpha_1^2 + \alpha_2^2 - (\alpha_1^2 - \alpha_2^2)^2$.  

\end{corollary} 

\vspace{4mm}

\section{\uppercase{Singularities and Projective Transformations}}
\label{sec:Recasting the P3P Problem}

The next goal will be to put equation (6) into a somewhat simpler form using a simple transformation. Towards this end, recall that the curve on the unit sphere given by (4) and $x^2 + y^2 + z^2 = 1$ is specifically determined by the angle $2 \theta_0$ between the two fixed great circles, which determines $\mu_0 = \cos \theta_0$ and $\nu_0 = \sin \theta_0$, and by the constant parameters $\alpha_1$ and $\alpha_2$, which determine the location of $\hat{a}$ with respect to $\hat{a}_1$ and $\hat{a}_2$. The cosine of the length of the sliding arc, $\hat{a}_1 \cdot \hat{a}_2$, which equals $(1 - \alpha_1^2 - \alpha_2^2) / (2 \alpha_1 \alpha_2)$, must remain constant. 

When the parameters are suitably chosen, it is possible to slide the arc into a position where it crosses the $z = 0$ great circle (the ``equator"). Assume that the parameters are set so that this point of intersection is $\hat{a}$. Now, the curve is clearly symmetric under reflecting about the $xy$-plane (sending $z$ to $-z$). This reflection would typically move the arc into a different position, but without moving $\hat{a}$. Therefore, this $\hat{a}$ would be a double point for the curve, a point where the curve crosses itself.  Let us now determine the coordinates of $\hat{a}$ in this situation. 

Using (2), it is straightforward to derive $x \, y \; = \; \mu_0 \nu_0 \, (\alpha_2^2 \, \nu_2^2 - \alpha_1^2 \, \nu_1^2)$. But, $\nu_1^2 = 1 - \mu_1^2$ and $\nu_2^2 = 1 - \mu_2^2$, so  $x \, y \; = \; \mu_0 \nu_0 \, (\alpha_2^2 - \alpha_1^2  - \alpha_2^2 \, \mu_2^2 + \alpha_1^2 \, \mu_1^2)$. Setting $z = 0$ in system (1) yields $\alpha_2 \, \mu_2 = - \alpha_1 \, \mu_1$, and so in this case,   

\vspace{-3mm}

$$2 \, x \, y \; = \; 2 \, (\alpha_2^2 - \alpha_1^2) \, \mu_0 \, \nu_0 \; = \; (\alpha_2^2 - \alpha_1^2) \, \sin 2\theta_0.$$ 

\vspace{2mm}

\noindent But of course $x^2 + y^2 = 1$, so we effectively have the intersection of a right hyperbola with the unit circle (in the $xy$-plane). There are no real solutions when $|(\alpha_2^2 - \alpha_1^2) \, \sin 2\theta_0| > 1$, but there are four symmetrically placed solutions when $|(\alpha_2^2 - \alpha_1^2) \, \sin 2\theta_0| < 1$. In the latter case, we find that 

%\vspace{-2mm}

$$ x^2, \, y^2 \; = \; \frac{1 \pm \sqrt{ \eta }}{2}, $$

%\vspace{2mm}

\noindent where 

\vspace{-2mm}

$$ \eta \; = \;  1 - 4 \, (\alpha_1^2 - \alpha_2^2)^2 \, \mu_0^2 \, \nu_0^2. $$

\vspace{2mm}

\noindent Assuming that $x > 0$ and $y > 0$, we find that the proportion $x : y$ equals 

%\vspace{-2mm}

$$\sqrt{ \; \frac{1 \pm \sqrt{ \eta } }{2}}  \; \; : \; \; \sqrt{ \; \frac{1 \mp \sqrt{ \eta } }{2}} \; \; \; =  $$

$$ 1 \pm \sqrt{ \eta } \; \; : \; \;  2 \, (\alpha_2^2 - \alpha_1^2) \, \mu_0 \, \nu_0.$$

\vspace{2mm}

\noindent These points on the sphere correspond to the singular points in the projective plane, for which 

$$X : Y : Z \; \; = \; \; 1 \pm \sqrt{ \eta } \; \; : \; \;  2 \, (\alpha_2^2 - \alpha_1^2) \, \mu_0 \, \nu_0 \; \; : \; \; 0.$$

\vspace{2mm}

\begin{proposition} 

The cosine of the angle between the two vectors $( \, 1 \pm \sqrt{ \eta } \; , \;  2 \, (\alpha_2^2 - \alpha_1^2) \, \mu_0 \, \nu_0 \; , \; 0 \, )$ in $\mathbb{R}^3$ is 
$\pm 2 \, (\alpha_2^2 - \alpha_1^2) \, \mu_0 \, \nu_0$. 

\end{proposition} 

\begin{proof} 
 The dot product of these two vectors is \linebreak $1 - \eta + 4 \, (\alpha_2^2 - \alpha_1^2)^2 \, \mu_0^2 \, \nu_0^2 $ \ = \  $8 \, (\alpha_2^2 - \alpha_1^2)^2 \, \mu_0^2 \, \nu_0^2$. The dot product of each of the two vector with itself is $1 \pm 2 \, \sqrt{\eta} + \eta  + 4 \, (\alpha_2^2 - \alpha_1^2) ^2 \, \mu_0^2 \, \nu_0^2 $ \ = \ $2 \, (1 \pm \sqrt{\eta})$. The product of these is $4 \, (1 - \eta) \; = \; 16 \, (\alpha_2^2 - \alpha_1^2) ^2 \, \mu_0^2 \, \nu_0^2 $. The square root of this is $| \, 4 \, (\alpha_2^2 - \alpha_1^2) \, \mu_0 \, \nu_0 \, |$. When the dot product of the two vectors is divided by the product of their lengths, the result is $\pm 2 \, (\alpha_2^2 - \alpha_1^2) \, \mu_0 \, \nu_0$. By basic properties of vectors, this equals the cosine of the angle between the two vectors. 
\end{proof} 

\noindent Even when there are no real points of the curve on the equator ($z = 0$), {\it i.e.} when $\eta <  0$, the constants may be extended to the field of complex numbers $\mathbb C$ to thereby obtain singularities for the corresponding complex curve in the complex projective plane $\mathbb{CP}_2$ . Also, with the aid of algebra manipulation software, it is straightforward to check that these points do indeed satisfy equation (6), and moreover, the gradient of the polynomial on the right side of (6) also vanishes at these points. This shows directly that these two points are singularities for the curve given by (6). 

Now, a simple projective transformation (a non-singular linear transformation of $\mathbb{R}^3 $ interpreted projectively in $\mathbb{RP}_2$) can be applied to (6) in order to move the singularities to the two points (at infinity) whose homogeneous coordinates are 1 : 0 : 0 and 0 : 1 : 0.  In fact, there is a whole family of such transformations, which will be parameterized here using scalars $\lambda_1$, $\lambda_2$ and $\lambda_3$.   Let $M$ be the following matrix: 

\begin{equation} 
\end{equation} 
$$\left[ \begin{array}{ccc} 
  \lambda_1 ( 1 - \sqrt{ \eta })  &   \lambda_2 ( 1 +\sqrt{ \eta })  & 0 \\ 
  2 \, \lambda_1 \, (\alpha_2^2 - \alpha_1^2) \, \mu_0 \, \nu_0  &   2 \, \lambda_2 \, (\alpha_2^2 - \alpha_1^2) \, \mu_0 \, \nu_0  & 0 \\ 
  0 & 0 &  \lambda_3
 \end{array} \right],$$  
 
\vspace{4mm}

 \noindent Now apply
 
\vspace{-2mm}

 \begin{equation} 
 \left[ \begin{array}{c} X \\ Y \\ Z  \end{array} \right] \; \; = \; \; \; M \;  \left[ \begin{array}{c} U \\ V \\ W  \end{array} \right]
 \end{equation} 

\vspace{4mm}

\noindent to express $X$, $Y$ and $Z$ in terms of $U$, $V$ and $W$, and substitute into equation (5) to achieve the next result. \\

\begin{corollary} 

Using the projective transformation given by (7) and (8), the curve given by equation (6) is transformed into the curve given by the following homogeneous equation: 

\vspace{-2mm}

\begin{equation}
\end{equation}
\vspace{-4mm}
$$ \begin{array}{lll}
0 & = & 16 \, (\alpha_1^2 - \alpha_2^2)^2 \; \mu_0^2 \; \eta^2 \; \lambda_1^2 \, \lambda_2^2  \; U^2 \, V^2 \\ \ \\
\ & + & 2 \, \big\{ \; \eta - 2 \, \beta \, \mu_0^2 \\
\ & \ & - \; [ \, 1 - 2(\alpha_1^2+\alpha_2^2) \, \mu_0^2 \, ] \; \sqrt{ \eta } \; \big\} \; \lambda_1^2 \, \lambda_3^2  \; U^2 \, W^2 \\ \ \\
\ & + & 2 \, \big\{ \; \eta - 2 \, \beta \, \mu_0^2 \\
\ & \ & + \; [ \, 1 - 2(\alpha_1^2+\alpha_2^2) \, \mu_0^2 \, ] \; \sqrt{ \eta } \; \big\} \; \lambda_2^2 \, \lambda_3^2  \; V^2 \, W^2 \\ \ \\
\ & - & 32 \,  (\alpha_1^2 - \alpha_2^2)^2 \, \beta \; \mu_0^4 \, \nu_0^2  \; \lambda_1 \, \lambda_2 \, \lambda_3^2  \; U \, V \, W^2 \\ \ \\ 
\ & + &  [ 1 - (\alpha_1+\alpha_2)^2 ]  [ 1 - (\alpha_1-\alpha_2)^2 ] \; \mu_0^2 \; \lambda_3^4  \; W^4. 
\end{array} $$

\end{corollary} 

\vspace{4mm}

As a polynomial in $U$, $V$ and $W$, notice that the right side has terms that only involve these monomials: $U^2 \, V^2$, $U^2 \, W^2$, $V^2 \, W^2$, $U \, V \, W^2$ and $W^4$. Next, we ask what happens to (6) and (9) in the special case where $\eta = 0$? To begin, notice that 

\vspace{-2mm}

 $$\eta = 0 \; \; \Longleftrightarrow \; \; \mu_0^2 = \frac{\alpha_1^2 - \alpha_2^2 \pm\sqrt{(\alpha_1^2 - \alpha_2^2)^2 - 1)}}{2(\alpha_1^2 - \alpha_2^2)} {\ \atop .} $$

\vspace{2mm}

\noindent Direct substitution into (9) leads to the polynomial equation 

\vspace{-2mm}

$$ \begin{array}{lll}
0 & = & W^2 \; \big\{ -4 \beta (\lambda_1 U + \lambda_2 V)^2 \; \; + \\
\ & \ & [ 1 - (\alpha_1+\alpha_2)^2 ] [ 1 - (\alpha_1-\alpha_2)^2 ] \; \lambda_3^2 \; W^2 \; \big\} 
\end{array} $$

\vspace{2mm}

\noindent when $\eta = 0$. This factorization is not mirrored in (6) however, which reduces to the following when $\eta = 0$: 

\vspace{-2mm}

$$ \begin{array}{lll}
0 & = &  (\alpha_1^2 - \alpha_2^2)^2 \; [ \; (X + Y)^4 + 4 \, X \, Y \, Z^2 \; ] \\
\ & + & 2 \, [ \;  2 \, (\alpha_1^2 - \alpha_2^2)^2 - \alpha_1^2 - \alpha_2^2 \; ] \; (X^2 + Y^2) \, Z^2 \\
\ & - & 2 \, (\alpha_1^2 - \alpha_2^2) \; \sqrt{ (\alpha_1^2 - \alpha_2^2)^2 - 1} \; (X^2 - Y^2) \, Z^2. 
\end{array} $$

\noindent Of course the matrix $M$ is singular when $\eta = 0$, so we no longer have a projective transformation between the curves given by (6) and (9). 

Another special case where $M$ is also singular is when $|\alpha_1| = |\alpha_2|$. This means that the point $\hat{a}$ is located midway between $\hat{a}_1$ and $\hat{a}_2$ along a great circle containing these three points. In this case, equation (9) becomes simply 0 = 

\vspace{-2mm}

$$W^2 \; [ \; 4 \,(1 - 4 \, \alpha_1^2 \, \mu_0^2) \; \lambda_2^2  \; V^2 \; + \; (1 - 4 \alpha_1^2) \, \mu_0^2  \, \lambda_3^2 \; W^2\; ].$$

\vspace{2mm}

\noindent The corresponding equation (6) becomes

\vspace{-2mm}

$$ \begin{array}{lll}
0 & = & X^2 \, Y^2 \, + \, (1 - 4 \alpha_1^2 \mu_0^2) \, \nu_0^2 \, X^2 \, Z^2 \\ \ \\ 
\ & + & (1-4 \alpha_1^2 \nu_0^2) \, \mu_0^2 \, Y^2 \, Z^2 + (1 - 4 \alpha_1^2) \, \mu_0^2 \, \nu_0^2 \, Z^4.
\end{array} $$

\vspace{2mm}

\noindent By setting $X = x$, $Y = y$ and $Z = \pm \sqrt{1-x^2-y^2}$ in either of the above special cases, the corresponding special case of (4) can be produced. 

We will henceforth assume that $\eta \ne 0$ and that  $|\alpha_1| \ne |\alpha_2|$, so that the two curve singularities are distinct, and so that the matrix $M$ is nonsingular. If one is actually interested in applying the linear transformation represented by the matrix $M$ in a situation where floating point roundoff error could be an issue, then it would be prudent to be concerned about the nature of the entries in $M$. Ideally, the three column vectors in $M$ ({\it i.e.} the images of the three standard vectors) should be roughly the same length and roughly orthogonal. Remember though that there is a whole family of possible choices for $M$, owing to the parameters $\lambda_1$, $\lambda_2$ and $\lambda_3$. 
These can be used to scale the column vectors to make them unit vectors. Proposition 2 then indicates that the smaller $| \, (\alpha_2^2-\alpha_1^2) \, \mu_0 \, \nu_0 \, |$ is, the closer these unit vectors will be to forming an orthonormal system. 

\section{\uppercase{Recasting the P3P Problem}}
\label{sec:Recasting the P3P Problem}

The starting point for the investigation reported on in this manuscript was the Perspective 3-Point (P3P) problem. This old and well-known problem was briefly outlined in the introduction. The objective now is to recast the P3P problem in terms of the arc-sliding problem that was just introduced. We here take a ``camera-centric" view of the P3P problem, meaning that attention will be limited to any given Cartesian coordinate system for which the camera's center of perspective $O$ is located at the origin, and for which the standard basis consists of vectors that are actually orthonormal in physical space. It is presumed that the directions to the three control points $P_1$, $P_2$, $P_3$ are known, but that the distances to these points are unknown. To avoid degenerate situations, it will tacitly be assumed that $O$, $P_1$, $P_2$ and $P_3$ are in general position; that is, they are the vertices of a non-degenerate tetrahedron. It is assumed also that the distances between the control points are known. The goal of course is to determine the distances from the origin to the control points, or at least produce a small set of possible distances. As soon as these distances become known, it becomes trivial to locate the control points in the camera-centric coordinate system.  

By recasting this problem in terms of the arc-sliding problem, it will be possible to produce a rather efficient and accurate algorithm to solve the P3P problem. In fact, this has been implemented and tested, as discussed in later sections. Moreover, the conversion of the P3P problem to the arc-sliding problem, and thence to a study of elliptic curves, is inherently interesting and holds the potential for further insights concerning the P3P problem. 

To begin the reformulation of the P3P problem, it will be helpful to introduce a little non-standard terminology. A line through the origin and one of the three control points will be called a ``view line." A plane containing two view lines will be called a ``containment plane." A ``sideline" will be any line containing two control points. Here we are regarding the control points as the vertices of a triangle, so that a sideline is the extension of one of the sides of the triangle. A ``translated sideline" will mean a line through the origin that is parallel to a sideline. Now, each line through the origin intersects the unit sphere in a unique pair of antipodal points that will be called the ``spherical points" for this line. Each plane through the origin intersects the unit sphere in a great circle that will be referred to as its great circle.

Here now are a few basic facts of critical importance to the reformulation of the P3P problem: 

\vspace{2mm}

\begin{enumerate}

\item There are six lines through the origin of interest here: the three view lines (known) and the three translated sidelines (unknown).

\item Each containment plane contains two view lines, plus the sideline between the two control points incident with these view lines, plus the corresponding translated sideline.

\item The great circle for each containment plane contains the spherical points for the two view lines and the translated sideline contained in that plane. 

\item Given two view lines, and one spherical point for each of these, the distance between these two points along the shorter great circle arc connecting them is equal to the one of the two (supplementary) angles between the view lines.

\item The cosine of this distance equals the dot product of the two vectors from the origin to the two spherical points. 

\item The intersection of two containment planes is a view line, and the corresponding two great circles intersect at the spherical points of this view line.

\item The angle between these two great circles (at either of the spherical points) equals the angle between the corresponding containment planes.

\item All three translated sidelines lie on a common plane, the plane through the origin that is parallel to the plane containing the control points.

\item The great circle for this plane contains the spherical points for the three translated sidelines. 

\item Given two translated sidelines, and one spherical point for each of these, the distance between these two points along the shorter great circle arc connecting them is equal to one of the two (supplementary) angles between the corresponding triangle sidelines. This is either the interior or exterior angle, at the control point where the sidelines meet, for the control-points triangle. 

\end{enumerate}

\begin{figure} 
\centerline{ \includegraphics[width=7cm]{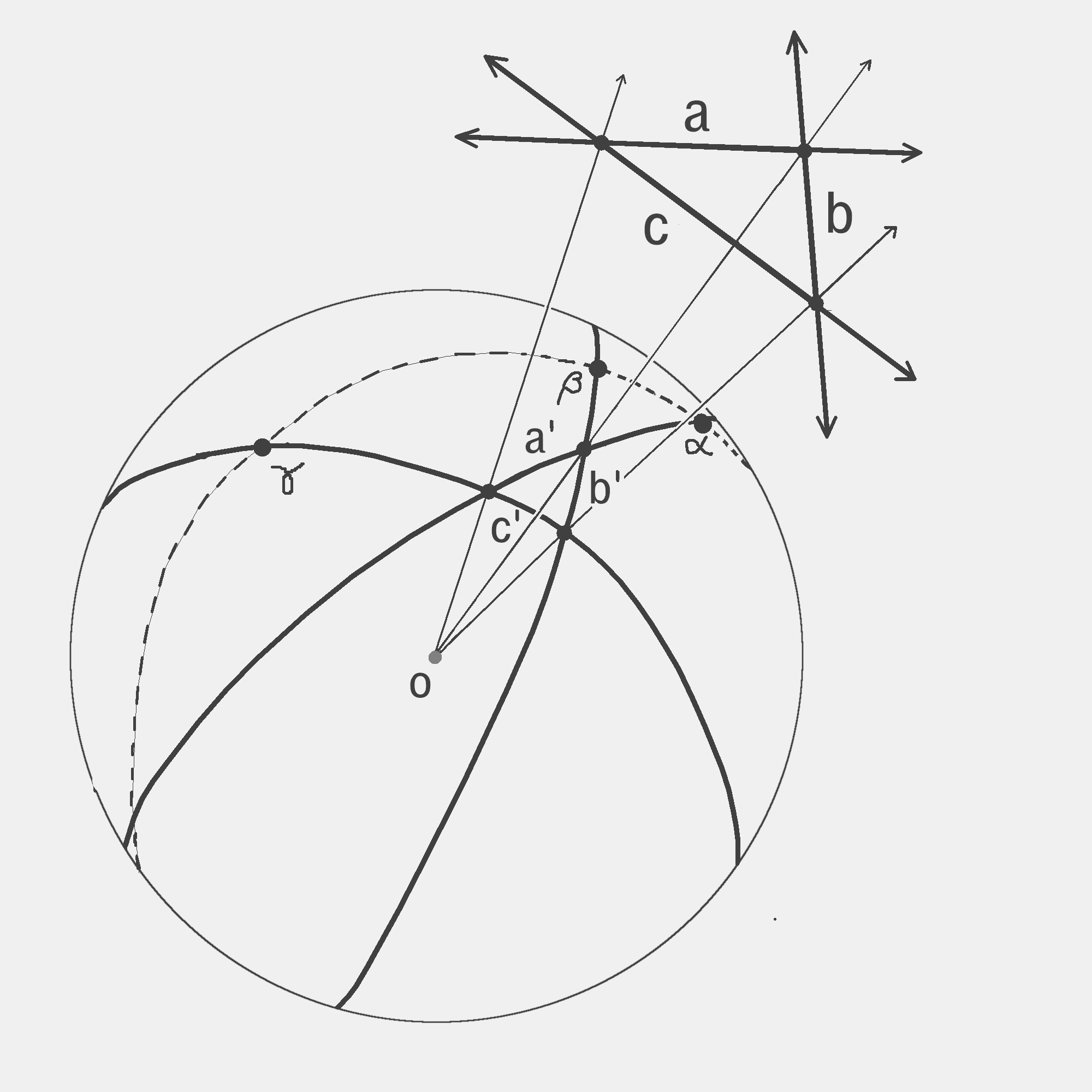}  }
\caption{projections onto the unit sphere} 
\label{fig:2}
\end{figure}

The three containment planes in the P3P problem now correspond to three known / fixed great circles on the unit sphere. The points where these circles intersect correspond to the view lines. Assuming for a moment that we actually knew the positions in space of the three control points, then we would also know the translated sidelines and the plane containing these. This plane would correspond to another great circle with three pairs of antipodal points, each pair representing a translated sideline. 

Now, even though this great circle and these points are not actually known, it is important to realize that the distances between these points along this great circle are known. This is so since the shape of the control-points triangle is known, and hence its interior and exterior angles are also known. So, a priori, we at least know how the three pairs of antipodal points corresponding to the translated sidelines are configured on the great circle that contains them. Let us call this configuration the ``desired configuration." Remember also that each spherical point corresponding to a translated sideline must also lie on a great circle corresponding a containment plane. 

The geometric features discussed above can largely be seen in Figure 2, where the control-points triangle has sidelines labeled $a$, $b$ and $c$, and the containment planes for these have corresponding great circles labeled $a'$, $b'$ and $c'$. The translated sidelines have corresponding spherical points labeled $\alpha$, $\beta$ and $\gamma$ (and their antipodes). The great circle containing these, whose plane is parallel to the control-points triangle plane, is drawn dashed. (The view presented here makes it perhaps difficult to believe that the line $\overleftrightarrow{O\alpha}$ is parallel to the sideline $\overleftrightarrow{a}$, etc., but this is intended to be the case.)

So, the recast problem now becomes clear. It is necessary to find a great circle that intersects three known / fixed great circles in such a way that the points of intersection achieve the desired configuration. When such a circle is found, it is a candidate for the great circle of the translated control-points plane. It then becomes straightforward to determine how to translate this plane through the origin so that its intersection with the view lines yields three points that achieve the known distances between the control points. The details of this procedure will be presented in the Section 6. In this way, either the control points are found, or else the points for some other P3P solution are found. 

Thus far we have only hinted at a relationship between the P3P problem and the arc-sliding problem, but have not yet made the complete connection. We will do so now. Finding a circle that has intersections with the three given circles that match the desired configuration can be handled in the following manner. Ignore one of the three fixed great circles for the moment.  (Which one to ignore is actually important in the algorithm in Chapter 6, but this issue will be ignored here.) Consider a moving great circle that always intersects the other two fixed great circles in accordance with the desired configuration. 

Now, focus on a pair of antipodal points on the moving circle, specifically the points that match the desired configuration. That is, these points are positioned on the moving circle where it {\it should} intersect the fixed great circle that we have been ignoring so far. Let us call these antipodal moving points, the ``tracing points." As the moving circle is slid around, the curve traced out by the tracing points is completely understood. This is because we have already solved the arc-sliding problem. 

To solve the P3P problem, we now want to know where this curve intersects the fixed great circle that we have been ignoring until now. The situation becomes even simpler when we transfer this problem from the unit sphere to the projective plane. That is, we can switch focus from the curve (4) on the sphere, to the corresponding curve (6) in the projective plane. In switching from the sphere to the plane, the various great circles are replaced with lines. Thus, the problem reduces to solving a system involving a fairly simple quartic equation, together with a linear equation. It is also possible to use the projective transformation from Section 3 to translate this problem into the problem of simultaneously solving equation (9) and a linear equation. Indeed, the successful P3P algorithm discussed in Section 6 does precisely this. The details are provided there.

\section{\uppercase{A Family of Quartic Curves}}
\label{sec:A Family of Quartic Curves}

Equation (9) involves a homogeneous quartic polynomial in three variables, which when de-homogenized by setting $u = U/W$ and $v = V/W$ takes on the following form: 
  
\vspace{-2mm}

\begin{equation} 
a^2 \; u^2 \, v^2 \; - \; b^2 \; u^2 \; - \; c^2 \; v^2 \; + \; d^2 \; - \; 2 \, e^2 \, u \, v.  
\end{equation} 

\vspace{2mm}

\noindent The coefficients here might seem a bit surprising since they involve squares and so forth, but these have been chosen to facilitate the analysis, as will be seen. In the arc-sliding problem, the coefficients were real, but potentially negative. So to make (9) match (10), it might be necessary to extend the field of constants to the complex number field. (Despite this, the algorithm in Section 6 avoids complex numbers.)  If one were to use (10) in the context of some other field of constants, it might be necessary to similarly extend the field of constants, in order to make sense of the analysis. The polynomial (10) will henceforth be referred to as $\mathcal{Q}$.  \\

\begin{proposition} 
A polynomial in two variables $p(u,v)$ has the form (10) (after possibly enlarging the field of constants) if and only if it is of degree no more than four, is invariant under negating $u$ and $v$ simultaneously, and has singularities at infinity along the $u$-axis and along the $v$-axis. \\
\end{proposition} 

\vspace{-2mm} 

\noindent To clarify the part about singularities, the technical meaning is that when $p(u,v)$ is homogenized to get $\tilde{p}(U,V,W)$, by setting $u = U/W$ and $v = V/W$, then $\tilde{p}(U,V,W)$ and its gradient $\vec{\nabla} \tilde{p} \, (U,V,W)$ both vanish at the points (1,0,0) and (0,1,0). 

\begin{proof}
The general quartic polynomial in $u$ and $v$ has a term for each of the following monomials: $u^4$, $u^3 v$, $u^2 v^2$, $u v^3$, $v^4$, $u^3$, $u^2 v$, $u v^2$, $v^3$, $u^2$, $u v$, $v^2$, $u$, $v$ and 1.  First, assume that $p(u,v)$ has all of the properties given after ``if and only if." Because of the invariance, it can only involve these monomials: $u^4$, $u^3 v$, $u^2 v^2$, $u v^3$, $v^4$, $u^2$, $u v$, $v^2$ and 1. Now consider the homogenized version $\tilde{p}(U,V,W)$, which can only involve following monomials:  $U^4$, $U^3 V$, $U^2 V^2$, $U V^3$, $V^4$, $U^2 W^2$, $U V W^2$, $V^2 W^2$ and $W^4$. Since $\tilde{p}(1,0,0) = 0 = \tilde{p}(0,1,0)$, the monomials $U^4$ and $V^4$ cannot occur (they must have zero coefficients). A straightforward check of the gradient condition further reveals that the monomials $U^3 V$ and $U V^3$ cannot occur either. This leaves the following monomials: $U^2 V^2$, $U^2 W^2$, $U V W^2$, $V^2 W^2$ and $W^4$. So $p(u,v)$ can only involve these monomials: $u^2 v^2$, $u^2$, $u v$, $v^2$ and 1, and is therefore of the form (10) (extending the field of constants if need be). This establishes the ``if" part of the claim. Based on the preceding reasoning, the ``only if" part is now trivial. 

\end{proof}

Besides the fact that (10) is a general form that encompasses the polynomials in (9), it is also a large enough family of polynomials to include some polynomials that have gained attention in the area of cryptography. The polynomials for Edwards curves \cite{E} and Twisted Edwards curves \cite{B} both match the form of (10), and the following analysis pertains to both of these families of curves. Indeed much of the motivation for and guidance in the present work came from these two cited papers.  \\ 

\begin{theorem} 
Letting $\mathcal{Q}$ denote the polynomial (10), in the two variable $u$ and $v$, and assuming that the constants $a$, $b$, $c$ and $d$ are nonzero, the curve given by $\mathcal{Q} = 0$ is birationally equivalent to the curve 
  
\vspace{-2mm}

\begin{equation}
\xi^2 \; \; = \; \;  ( \; 1 - \omega^2 \; ) ( \; 1 - \kappa^2 \; \omega^2 \; ), 
\end{equation} 
  
\vspace{2mm}

\noindent where $\omega$ and $\xi$ are variables with $\omega \; = \; u \; / \sqrt{\rho} $ \ and 
  
\vspace{-2mm}

$$\xi \; = \; \frac{(a^2  \, u^2 \, - \, c^2) \, v \, - \, e^2 \, u}{c \, d},  $$ 
 
\vspace{2mm}

\noindent and where some needed constants are defined are follows: 
  
\vspace{-2mm}

$$ \delta \; = \; (a d + b c + e^2)(a d + b c - e^2) \cdot $$ 
$$ (a d - b c + e^2)(a d - b c - e^2), $$ 

$$\rho \; = \;  \frac{a^2 d^2 + b^2 c^2 - e^4 \, + \,  \sqrt{\delta}}{2 a^2 b^2}$$ 

$$\hbox{and} \quad \kappa  \; = \;  \frac{a b}{c d} \, \rho \, {\ \atop .} $$ 

\vspace{2mm}

\noindent (The square root symbols here represent arbitrary square roots, and again, the field of constants may need to be extended to accommodate these constants.) 

\end{theorem} 

\begin{proof} 

Upon rearranging the terms, we see that $\mathcal{Q} \; = \; ( a^2 \, u^2 - c^2 ) \, v^2  - 2 \, e^2 \, u \, v + (d^2 - b^2  \, u^2).$ So,  $(a^2 \, u^2 - c^2) \mathcal{Q} \; = \; [ \; ( a^2 \, u^2 - c^2 ) v -  e^2  u \; ]^2 \, - \, e^4 u^2 \, + \, (a^2 u^2 - c^2)(d^2  -  b^2 u^2)$, and $(a^2 u^2 - c^2) \, \mathcal{Q}  \, /  \, (c^2 d^2) \; = \; \{ \, [ \, ( a^2 \, u^2 - c^2 ) v \, - \, e^2  u \, ] \; / \; (c d) \, \}^2 \;- \; \{ 1 + f \, u^2 + g \, u^4 \}$,  where \  $f \; = \; ( \, e^4 - \, a^2 \, d^2 \, - \, b^2 \, c^2) \; / \; (c^2 \, d^2)$ \ \ and \ \ $g \; = \; (a^2 \, b^2) \, / \, (c^2 \, d^2).$ 

Next notice that $(1 + f \, \rho + g \, \rho^2) = 0$. (Algebraic manipulation software can help here). So, if $\mathcal{Q} \; = \; 0$, then $\xi^2 \; = \; 1 + f \, u^2 + g \, u^4  \; = \; 1 + f \, \rho \, \omega^2 + g \, \rho^2 \, \omega^4$. The right side of this equation is a quadratic polynomial in $\omega^2$, and is zero when $\omega^2 \; = \; 1$. It therefore factors as $(1 - \omega^2)(1 - \lambda \, \omega^2)$, for some number $\lambda$. 

By comparing coefficients of $\omega^4$, it becomes clear that  $\lambda \; = \; g \, \rho^2  \; = \; \kappa^2$. The transformation from $(u, v)$ to $(\omega, \xi)$ is clearly rational. It is straightforward to invert this and find that the inverse transformation is also rational. 

\end{proof} 

\begin{corollary} 
The curve $\mathcal{Q} \; = \; 0$ is generally a genus-one quartic curve. It is therefore (generally) birationally equivalent to some non-singular cubic curve. As such, its $j$-invariant equals 
  
\vspace{-2mm}

$$ \frac{16 \, (a^4 b^4 \rho^4 + 14 a^2 b^2 c^2 d^2 \rho^2 + c^4 d^4)^3}{a^2 b^2 c^2 d^2 \rho^2 (a^2 b^2 \rho^2 - c^2 d^2)^4} \; = $$
  
$$ \frac{16 \, \left( \begin{array}{c} 
a^4 d^4 \, + \, b^4 c^4 \, + \, e^8 \, + \, 14 a^2 b^2 c^2 d^2 \\
- \; \; 2 a^2 d^2 e^2 \; \; - \; \; 2 b^2 c^2 e^4 \,  \end{array} \right)^3}{a^2 \, b^2 \, c^2 \, d^2 \; \delta^2} {\ \atop .} $$
  
\vspace{2mm}

\end{corollary} 

\begin{proof} 

$\mathcal{Q} \; = \; 0$ generally describes a quartic curve, and is birationally equivalent to a Jacobi curve (non-degenerate generally) given by the equation (11). The latter has genus one and is birationally equivalent to some non-singular cubic curve, so the same can be said of the former. Moreover, since the $\mathcal{Q} \; = \; 0$ curve has degree four and two singularities (as indicated in Proposition 1), the Genus-Degree Theorem also ensures that its (geometric) genus is at most $(4-1)(4-2)/2 - 2 \cdot (2)(2-1)/2 \; = \; 3 - 2 = 1$. Since the curve is generally not rational, it must generally have genus one.  

The $j$-invarient of a genus-one curve is invariant under birational transformations, so the $\mathcal{Q} \; = \; 0$ curve has the same $j$-invariant as the Jacobi curve. For the latter, this number is known to equal 
$$16 \, (\kappa^4 + 14 \kappa^2 + 1)^3 \; / \; [ \, \kappa^2 (1-\kappa^2)^4 \, ].$$ 
The first formula for the $j$-invariant in the corollary is then obtained by substituting $a b \, \rho \, / \, c d$ for $\kappa$ in this formula. The second formula requires some manipulations using the equations concerning the constants in the theorem.   

\end{proof} 

The curve described by equation (11) is a famous quartic curve associated with Jacobi's development of elliptic functions. It is well-known that it can be parameterized by letting $\omega \; = \; \hbox{sn } t$ and $\xi \; = \; \hbox{cn } t \hbox{ dn } t$, where sn, cn and dn are Jacobi elliptic functions. Less known is the fact that other parameterizations using elliptic functions are possible. For instance, letting $\omega \; = \; \hbox{cd } t$ and $\xi \; = \; (1 - \kappa^2) \hbox{ sd } t \hbox{ nd } t$. Another is obtained by using $\omega \; = \; (1/\kappa) \hbox{ ns } t$ and $\xi \; = \; (1/\kappa) \hbox{ ds } t \hbox{ cs } t$. Yet another results from setting  $\omega \; = \; (1/\kappa) \hbox{ dc } t$ and $\xi \; = \; [ \, (1-\kappa^2)/\kappa \, ] \hbox{ nc } t \hbox{ sc } t$. Any of these can be used to parameterize the Jacobi curve (11), and thence, via the birational transformation, to parameterize the curve whose equation is $\mathcal{Q} \; = \; 0$. Some experimentation with this has demonstrated that the different parameterizations combine to cover different portions of the $\mathcal{Q} \; = \; 0$ curve. These ideas concerning elliptic functions will not be exploited in the algorithm to be presented in the next section, and nothing further will be said about them in this paper. 

While there are some differences among authors concerning the terminology ``elliptic curve," and the curves $\mathcal{Q} \; = \; 0$ qualify by some standards, it can at least flatly be stated that these curves are birationally equivalent to non-singular cubic curves. As an example of this, the Jacobi curve can be birationally transformed into a curve studied by Legendre:  $\nu^2 \; = \; \mu (\mu - 1)(\mu - \lambda)$, where $\mu$ and $\nu$ are variables, and $\lambda \; = \; [ \, (1+\kappa) / (1-\kappa) \, ]^2$. The author has explored transforming the $\mathcal{Q} \; = \; 0$ curve to the Jacobi curve, and thence to the Legendre curve and to the Weierstrass curve, but nothing further will be reported about this here.

\section{\uppercase{A New P3P Algorithm and Implementation}}
\label{sec:A New P3P Algorithm}

The algorithm that will be introduced now has been implemented as a C program. All of the C code is publicly available at

\vspace{2mm} 

\centerline{\small \tt https://github.com/mqrieck/EllipticCurveP3P}

\vspace{2mm} 

\noindent The function named  {\tt ellipticP3PSolver} is the starting point for the implementation of this algorithm.  We are involved here with the classical P3P problem, where a pinhole camera is assumed, having known characteristics. It follows, and we take it for granted, that the directions (in the camera's frame of reference) to the three control points are easily discerned from the camera image. The directions are specified by three unit vectors. The distances between the control points are also presumed to be known, a priori. A P3P-solving algorithm receives this information as input, and returns as output the possible distances to the control points. This at least is how we will regard the problem.  In order to facilitate the discussion, it might be helpful to consider the algorithm and its C implementation as consisting of five phases, as suggested by the commenting in the C code. 

\subsection{Phase one}

 The C function {\tt ellipticP3PSolver} actually receives as input, three vectors $\vec{v}_1$, $\vec{v}_2$ and $\vec{v}_3$, in the form of a 3-by-3 matrix, that give the positions of the control points (in the camera's frame of reference).  This is done because this function not only implements the new P3P-solving algorithm, but also checks on the accuracy of its computations, for testing purposes. 
 
 It computes the differences between the input vectors, producing three other vectors, $\vec{s}_1$, $\vec{s}_2$ and $\vec{s}_3$, that represent the displacements between the control points. It computes the lengths of the $v$ vectors and the $s$ vectors. It produces normalized versions $\hat{v}_1$, $\hat{v}_2$ and $\hat{v}_3$ of the $v$ vectors, as well as normalized versions $\hat{s}_1$, $\hat{s}_2$ and $\hat{s}_3$ of the $s$ vectors. It computes the dot products between the normalized versions of the $v$ vectors, as well as the dot products between the normalized versions of the $s$ vectors. 
 
The original vectors ($\vec{v}_1$, $\vec{v}_2$ and $\vec{v}_3$) are only seen again at the end of {\tt ellipticP3Psolver}, in Phase Five, where they are used for the sole purpose of testing how well the algorithm did in locating the control points. In this way, most of the {\tt ellipticP3Psolver} code is an honest P3P solver, but a few lines of code at the end serve to analyze the execution of this algorithm. 

The algorithm is permitted to use the unit vectors $\hat{v}_1$, $\hat{v}_2$ and $\hat{v}_3$, which will henceforth be referred to as the ``(unit) view" vectors. The algorithm is also allowed to use the lengths of the vectors $\vec{s}_1$, $\vec{s}_2$ and $\vec{s}_3$, since these length are just the distances between the control points. The unit vectors $\hat{s}_1$, $\hat{s}_2$ and $\hat{s}_3$ will be referred to as the ``unit side vectors." The dot products between the unit side vectors may also be used in the algorithm since these numbers are just the cosines of the exterior angles for the triangle whose vertices are the control points, and this information is easily computed from the distances between the control points. 

Unit vectors  $\hat{c}_1$, $\hat{c}_2$ and $\hat{c}_3$ pointing perpendicular to the containment planes may also be used in the algorithm. This is legitimate since they can be computed from the unit view vectors $\hat{v}_1$, $\hat{v}_2$ and $\hat{v}_3$ by taking cross products and then rescaling these to get unit vectors. Let us call  $\hat{c}_1$, $\hat{c}_2$ and $\hat{c}_3$ the ``(unit) containment plane normals."  These unit vectors and the dot products between them are important in the new P3P-solving algorithm, so they are computed in Phase 1. Note that these dot products are the cosines of the (dihedral) angles between the containment planes. 

\subsection{Phase two}

This phase is chiefly concerned with selecting a rotation that moves one of the three view vectors so as to point vertically up, along the positive $z$-axis, while also moving the two containment planes that contain this vector so as to also become vertical and placed symmetrically about the $x$-axis. The two resulting containment planes will then intersect the unit sphere to yield two great circles that are aligned vertically, as in the setup from Section 2 and as seen in Figure 1. 

Of particular practical importance is the choice of which of the three view vectors to rotate into the vertical position.  Using the notation of Section 2, each of the three possible selections would result in a different value for $\theta_0$, and so corresponding values for $\mu_0$ and $\nu_0$. Since $\theta_0$ is half the (dihedral) angle between the two vertical containment planes, and since this angle is the same as the angle between their normals, the values of  $\mu_0$ and $\nu_0$ can be obtained rapidly from the geometric information gathered in Phase One. ($\theta_0$ won't actually be needed, but $\mu_0$ and $\nu_0$ will.)  

In addition, different choices for the rotation lead to different values for $\alpha_1$ and $\alpha_2$ that match the recast P3P problem, as discussed in Section 4. As a reminder, the interior and exterior angles between the sidelines of the control-points triangle determine a configuration of three pairs of antipodal points on a great circle. The configuration is known in the sense that the distances between these points along the circle can be easily calculated, as follows. 

One of the pairs of points is constrained to lie on one of the two vertical great circles and another pair is constrained to lie on the other vertical great circle. Letting $\pm \hat{a}_1$ be the points of the first pair, $\pm \hat{a}_2$ be the points of the second pair, and $\pm \hat{a}$ be the points of the remaining pair, the numbers $\pm \hat{a}_1 \cdot \hat{a}_2$, $\pm \hat{a}_1 \cdot \hat{a}$ and $\pm \hat{a}_2 \cdot \hat{a}$ are just the cosines of the interior and exterior angles of the control-points triangle. They are therefore known and usable in the algorithm. Moreover, Proposition 1 shows how to obtain the alphas from these dot products. 

The different rotation possibilities also leads to different values for $\eta \; ( \; = \;  1 - 4 \, (\alpha_1^2 - \alpha_2^2)^2 \, \mu_0^2 \, \nu_0^2  \; ).$ To avoid becoming involved with complex numbers, it is desirable to select a rotation that will prevent $\eta$ from being negative. Additionally, the projective transformation from Section 3 is required, and to minimize the distortion caused by this, it is desirable to keep the angle between the two singularities close to a right angle. By Proposition 2, this means keeping $| \, (\alpha_2^2 - \alpha_1^2) \, \mu_0 \, \nu_0 \, |$ as small as possible, except that due to a division issue in later computations, it should not be allowed to be too close to zero. For the same reason, none of the following should be too small: \ $| \mu_0 |$, $| \nu_0 |$, $| \alpha_1 |$, $| \alpha_2 |$. Based on these criteria, particularly the one about minimizing $| \, (\alpha_2^2 - \alpha_1^2) \, \mu_0 \, \nu_0 \, |$, it becomes possible to make an informed choice as to which view line would be best to rotate vertically. The evidence from extensive testing, presented in the next section, demonstrates that under reasonable circumstances, it is quite rare (if ever) that all of the choices are bad.   

\subsection{Phase three}

During Phase Three, the singularities, the (deforming) projective transformation matrix, its inverse, the rotation matrix and its inverse are all computed. The images of some needed vectors under the rotation and/or the deforming transformation are also computed. These computations are all straightforward based on the computations in the preceding phases and the analysis in the Section 3. The rotation matrix is efficiently computed using an algorithm that determines how to rotate two prescribed unit vectors to two other prescribed unit vectors, provided that the angles between the pairs of vectors are the same. This usually involves just computing a cross product and then constructing an Euler-Rodriguez matrix. However, in the rare case where the cross project is too small, a random rotation is applied to eliminate this obstacle. 

\subsection{Phase four}

This phase sets up a system of two equations in $U$ and $V$. These variables are the $U$ and $V$ in Section 3, except that we opt here not to work with homogeneous equations, and so set $W = 1$. The first of these is the equation in Corollary 2 (with $W = 1$). The second equation is a linear equation. It is obtained via the projective transformation of the linear equation in $X$, $Y$ and $Z$ that describes the line in the projective plane that corresponds to a great circle on the unit sphere. This great circle is the intersection of the unit sphere with one of the rotated containment planes, specifically, the containment plane that does not contain the view line that was rotated vertically. 

The system of equations in $U$ and $V$ is solved. This is accomplished by eliminating one of the variables to obtain a quartic equation in the other variable. The real roots of this equation are computed. This is done by first finding a real root of the cubic resultant of the quartic. This real root of a cubic polynomial must be computed with a high degree of precision. 

Each solution pair for $U$ and $V$ is transformed via the projective transformation to a solution triple for $X$, $Y$ and $Y$. Each of these solutions, when normalized, yields a point $\hat{a} \; = \; (x, y, z)$ on the unit sphere.

\subsection{Phase five}

For each of the solution points $\hat{a} \, = \, (x, y, z)$,  the corresponding values of $\mu_1$, $\nu_1$, $\mu_2$ and $\nu_2$ are computed using (2) and (5). These yield the coordinates of the constrained points $\hat{a}_1$ and $\hat{a}_2$, on the two vertical great circles. The great circle that passes through both of these (and also through $\hat{a}$) is then considered. A unit vector normal to the plane containing this circle is computed, and then rotated via the inverse of the original rotation. The result is a unit vector $\hat{n}$ that is normal to a plane through the origin that is potentially parallel to the plane containing the three control points, and that is at least parallel to the plane containing the points that constitute one of the solutions to the P3P problem. In the latter case, these points will also lie along the ``view lines," in the directions of the view vectors, and they will form the vertices of a triangle that is congruent to the control-points triangle.  (Actually, some care needs to be taken here to ensure that $\hat{n}$ is the normal that points towards the control points rather than away from them.) 

Some straightforward geometric / trigonometric reasoning is then applied to determine how much the plane would need to be translated in order to arrive at the control-points plane or the plane for some other P3P solution. For an indeterminate translation amount $\lambda > 0$, the distances between the points of intersection of the translated plane and each of the three view lines can be computed. For instance, using the first two view lines, this distance is $\lambda \; || \hat{v}_1 / ( \hat{v}_1 \cdot  \hat{n} ) -  \hat{v}_2 / ( \hat{v}_2 \cdot \hat{n} ) ||$, the square of which equals $\lambda^2 \; [ \;  ( \hat{v}_1 \cdot  \hat{n} )^2  +  ( \hat{v}_2 \cdot  \hat{n} )^2  -  2  ( \hat{v}_1 \cdot  \hat{n} )( \hat{v}_2 \cdot  \hat{n} )( \hat{v}_1 \cdot  \hat{v}_2 )    \; ] \; / \; [ \;  ( \hat{v}_1 \cdot  \hat{n} )^2 \,  ( \hat{v}_2 \cdot  \hat{n} )^2 \; ].$ Setting this equal to the square of the distance between the first two control points, we are able to determine the translation amount $\lambda$ that would allow these two intersection points to be the first two control points. 

Doing likewise with the other two pairs of view lines yields, in theory, the same translation amounts. In practice though, they are slightly different, due solely to roundoff error. Therefore, the three computed translation amounts are averaged, and this is taken to be a good approximation for the correct value of $\lambda$. The computed estimates for the distances to the three control points, or at least to three points that also solve the P3P problem, are then just $\lambda /  ( \hat{v}_1 \cdot  \hat{n} )$,  $\lambda /  ( \hat{v}_2 \cdot  \hat{n} )$ and  $\lambda /  ( \hat{v}_3 \cdot  \hat{n} )$, 

In this way, we have obtained one of the mathematical solutions to the P3P problem.  The {\tt ellipticP3Psolver} function now tests to see how close this triangle is to the control-points triangle. It uses a metric described in the next section for this. {\tt ellipticP3Psolver} returns the smallest of all these values among all of the P3P problem solutions it discovers.

\section{\uppercase{Experiments and Results}}
\label{sec:Experimental Results}

All of the experimentation reported about in this section was conducted using the C program mentioned in the preceding section, though some preliminary testing was also conducted using Python and Mathematica. The C program was run on a MacBook Pro with a 2.3 GHx Intel Core i5 processor. It was compiled using the clang-1001.0.46.4 compiler, The operating system was Mac OS 10.14.5.  During testing, an effort was made to prevent other applications from executing. As seen in the C source code, the timing results were obtained using the standard C {\tt time} library, and times were recorded from the start of a test trial until its completion. In tests that relied on multiple trials to obtain accurate statistics, these times were tallied and the total combined times were reported. 

As mentioned in the introduction, both the Lambda Twist and new elliptic-curve P3P solvers rely upon the accurate computation of a single real root of a cubic polynomial. The original C++ code for Lambda Twist \cite{PN} relied on the classical numerical method of Isaac Newton and James Raphson for this purpose. In contrast, the original C code implementation of the elliptic-curve method used a more algebraic approach, leveraging the classical methods of Gerolamo Cardano, Fran\c ois Vi\`{e}ta and Lodovico Ferrari. For comparison purposes though it seemed necessary that the two algorithms should share a common routine for extracting the cubic polynomial root. Both methods seemed to work better when the numerical approach was used, so this was substituted for the algebraic approach in the elliptic-curve P3P solver. 

Both algorithms avoid the need for complex numbers, and both lean substantially on linear algebra and require fast and accurate routines for floating-point matrix and vector manipulations. This was handled here using singly and doubly indexed arrays in the C language, mostly via simple C functions. A specialized function needed for the Lambda Twist method to compute eigenvalues and eigenvectors was written in C by adapting a similar function from the original Lambda Twist C++ code. The elliptic-curve method also has a special requirement for constructing a rotation matrix, and so a couple function were written for this purpose. 

All of the C code for implementing and testing the two P3P-solving algorithms was put into a single source file. This includes, for each of the two P3P-solving algorithms, a P3P-solving/checking function that receives as input, the presumed coordinates (in the camera's frame of reference) of the three control points. These functions use this information to compute the unit vectors that point in the directions of the control points, as well as the distances between the control points. Using only these unit ``view" vectors and distances between control points, each of these two functions goes on to apply its algorithm to estimate potential locations of the control points. At the end, the function compares its solutions, {\it i.e.} its potential control point locations, with the actual control point positions to see how well it did. 

For each solution, and each control point, the squared distance from the control point to the corresponding computed potential control point is calculated, and divided by the squared distance from the origin (the camera's center of perspective) to the control point. This then is the square of a relative error. The three squared relative errors for the three control points are added together, and the square root of this sum is then used as a measurement of the overall relative error for this particular solution. Among the solutions, the one with the least error is regarded as the ``correct" solution, and its error is returned by the P3P-solving/checking function and is regarded as the error associated with this particular trial of the P3P solver.  However, the P3P-solving/checking functions are rigged so that if some insurmountable obstacle arises during the application of the P3P-solving algorithm, a negative (sentinel) value is returned instead of the least error among the computed solutions. 

A student assistant (see acknowledgement) was employed to write Python code to thoroughly test and compare the two P3P-solving/checking (C) functions. These tests were subsequently rewritten in C and incorporated into the single C source code file. The tests were rerun in C and were found to be in very good agreement with the Python versions. Each test/experiment entailed calling one of the two P3P-solving functions, using certain fixed parameters, ten million times. 

\begin{figure} 
\centerline{ \includegraphics[height=5.2cm]{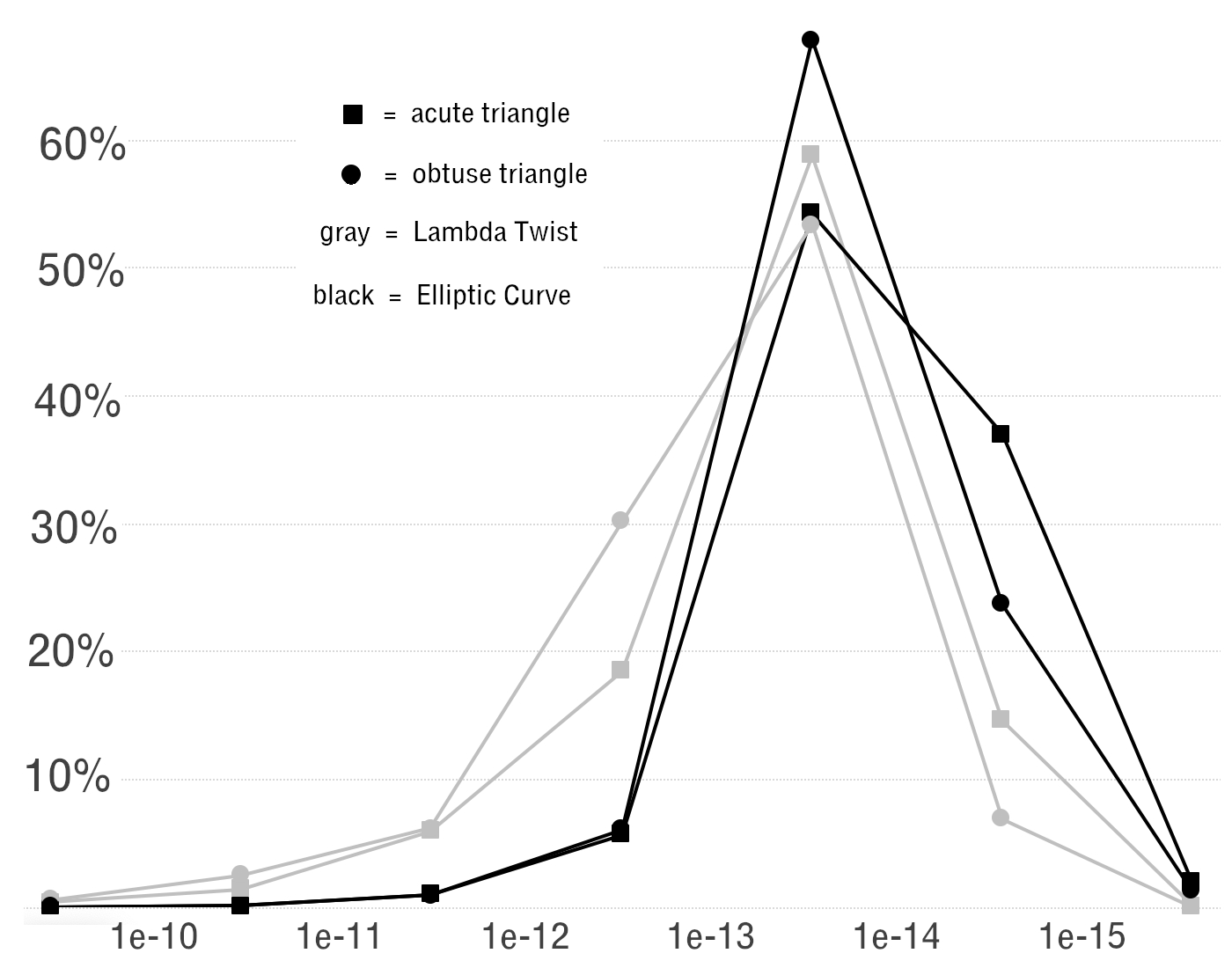}  }
\caption{attack angle range 0 - 30, \ lift range 10 - 20} 
\label{fig:3}
\end{figure} 

\begin{figure} 
\centerline{ \includegraphics[height=5.2cm]{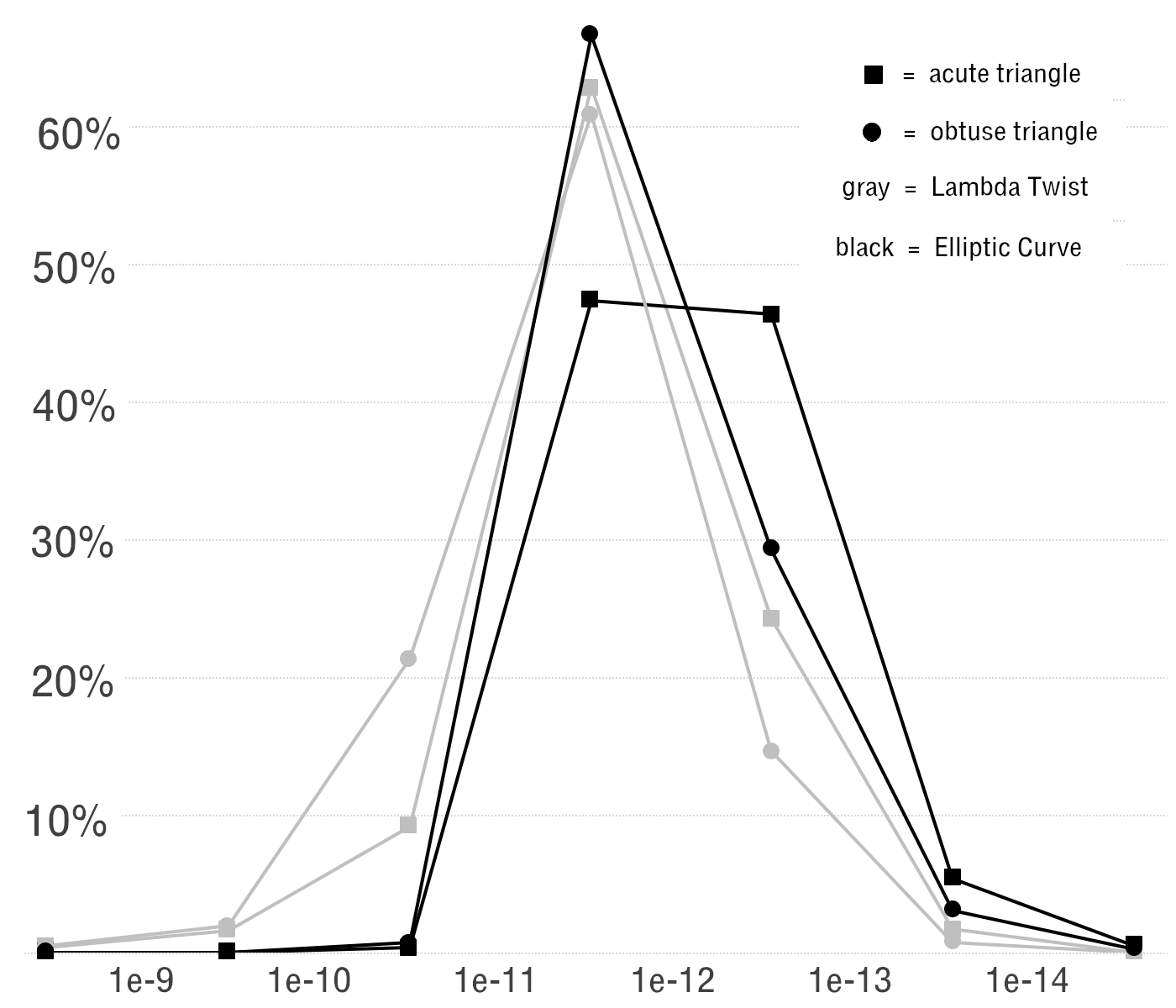}  }
\caption{attack angle range 0 - 30, \ lift range 100 - 200} 
\label{fig:3}
\end{figure} 

\begin{figure} 
\centerline{ \includegraphics[height=5.2cm]{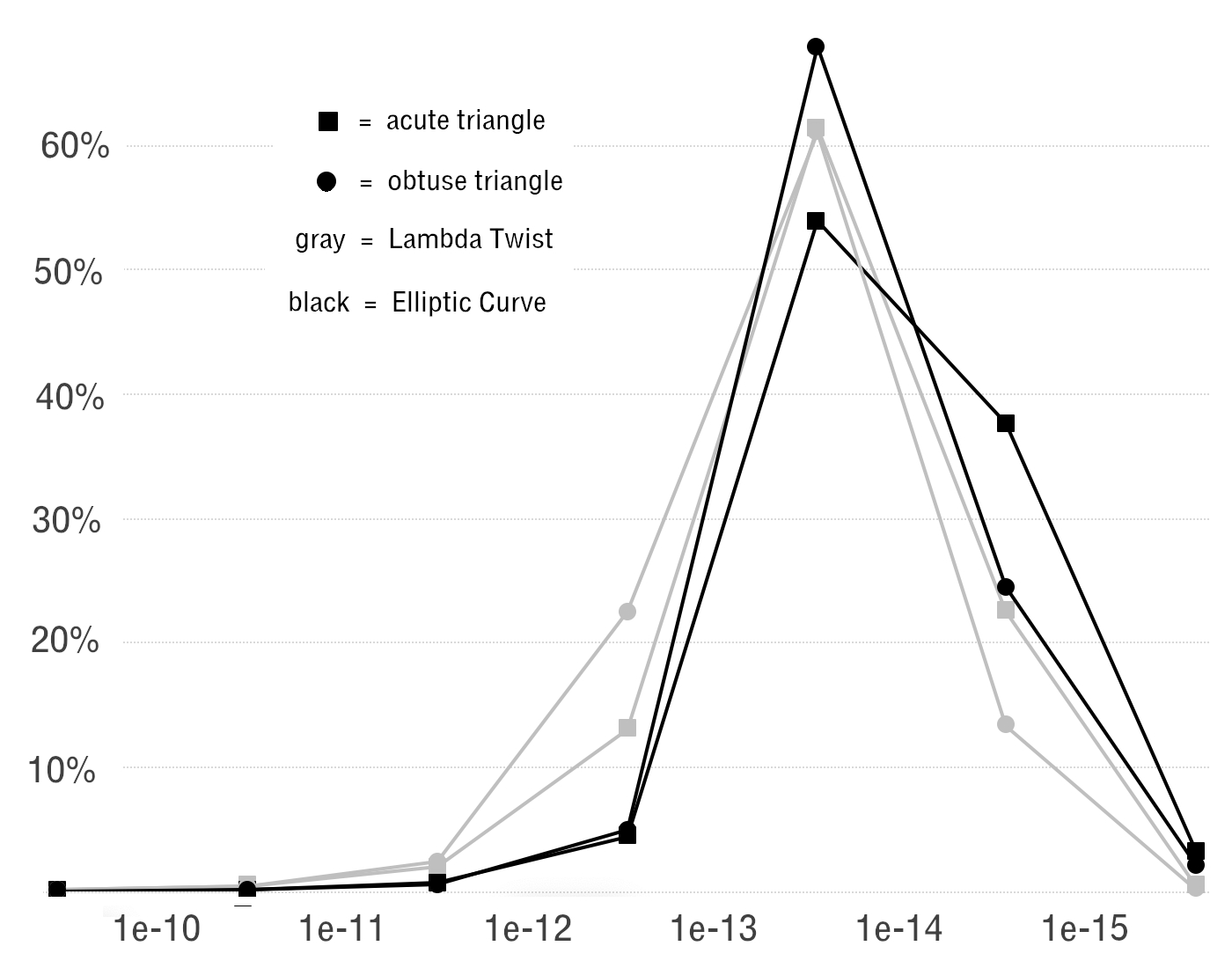}  }
\caption{attack angle range 30 - 60, \ lift range 10 - 20} 
\label{fig:4}
\end{figure} 

\begin{figure} 
\centerline{ \includegraphics[height=5.2cm]{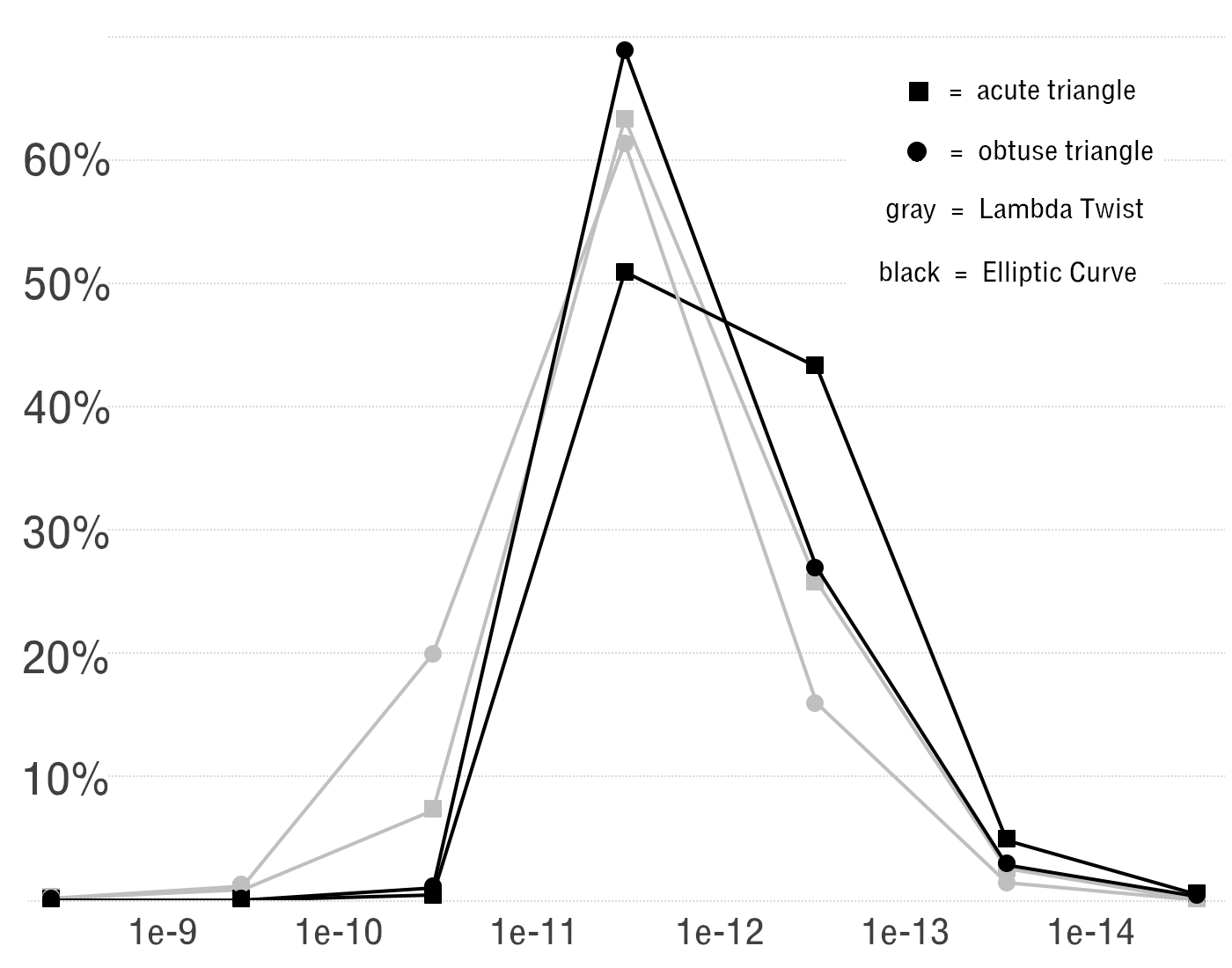}  }
\caption{attack angle range 30 - 60, \ lift range 100 - 200} 
\label{fig:5}
\end{figure} 

The program tallied the number of times the P3P-solving function succeeded in returning a valid (non-negative) minimal error number, and the number of times it failed and so returned a negative (sentinel) number. Failures were extremely rare (not more than four out of a million). As a technical point, this provides solid evidence that it was always possible or nearly always possible for Phase 2 of the elliptic curve method to make a good choice as to which of the view lines to rotate vertically.  

The tests also separated the valid returned error numbers according to the following numeric ranges:  0.01 to 0.1, 0.001 to 0.01, ..., 1e-14 to 1e-15. It tallied the number of returned error numbers that fell into each of these ranges. It also computed the average and standard deviation for the returned error numbers, overall. It also computed the minimum and maximum of the returned error numbers. It also computed the total CPU time for the combined trials. 

Early on, it was noticed that the reliability of the new elliptic-curve P3P method depends on the ``angle of attack," with the accuracy greatly diminished when this angle is too close to a right angle. Still, as the test data reveal, it does quite a good job when this angle does not exceed say sixty degrees. Consider the vector from the origin to the circumcenter of the control-points triangle. Also consider the vector from the origin to the nearest point on the control-points plane (which points in a direction orthogonal to this plane). The angle between these two vectors is what will be meant here by the phrase ``angle of attack" (or ``attack angle").  Because of the performance dependency, it was decided to conduct experiments for various ranges of attack angles. 

Each test/experiment was based on a prescribed triangle with vertices on the unit circle in the $xy$-plane. Each test proceeded as follows. Beginning with the prescribed triangle, a three-dimensional rotation was performed about a randomly selected axis in the $xy$-plane (through the origin), by a random amount. However, this amount was chosen so that when the triangle was next translated along the positive $z$-axis, the angle of attack fell somewhere in the range of attack angles allowed for the particular test being conducted. After this rotation, the triangle was translated (``lifted / elevated") by a random amount, by adding this amount to the $z$-coordinate of each triangle vertex. The amount of elevation was randomly chosen in some prescribed range. Then, another random rotation was applied, again rotating about a random axis in the $xy$-plane (through the origin), but this time by a random amount between -90 degrees and 90 degrees. 

The triangle vertices after applying the combined rigid motions then served as the control points for one trial of the test, and of course, this control-points triangle was always congruent to the original triangle.  Notice too that neither the amount of the lift nor the second rotation have any effect on the attack angle. So, only the first rotation determined the attack angle. 

Each test/experiment was the result of multiple trials of this sort (ten million per experiment), but all of the trials for a particular test used the same algorithm, the same starting triangle, the same range of possible rotation amounts for the first rotation, and the same range of possible lift amounts.  While many triangles were investigated, our systematic data collection focused on just two starting triangles, a typical acute triangle and a typical obtuse triangle. The $xy$-coordinates of the acute triangle were (0, 1), ($\cos 80^\circ$, $\sin 80^\circ$) and ($\cos 230^\circ$, $\sin 230^\circ$). The $xy$-coordinates of the obtuse triangle were (0, 1), ($\cos 70^\circ$, $\sin 70^\circ$) and ($\cos 300^\circ$, $\sin 300^\circ$). Two ranges for the random lift were investigated, from 10 to 20, and from 100 to 200. Also, two ranges for the attack angle were investigated, from 0 to 30 degrees, and from 30 to 60 degrees. \\ \ \\

\centerline{\small \sc Table 1} 
\vspace{-4mm} 
$$\begin{array}{cccc|ccccc} 
 \hspace{-2mm} \hbox{\tiny attack} & \hspace{-3mm} \hbox{\tiny lift} & \hspace{-4mm} \hbox{\tiny triangle} &  \hspace{-3mm}  \hbox{\tiny method} \hspace{-1mm} & \hspace{-1mm} \hbox{\tiny avg err} & \hspace{-1mm} \hbox{\tiny std dev} & \hspace{-1mm} \hbox{\tiny min err} & \hspace{-1mm} \hbox{\tiny max err} & \hspace{-1mm} \hbox{\tiny time} \\
\hline
 \hspace{-2mm} \hbox{\tiny 0-30} & \hspace{-3mm} \hbox{\tiny 10-20} & \hspace{-3mm} \hbox{\tiny \ acute} & \hspace{-3mm} \hbox{\tiny LT} & \hspace{-1mm} \hbox{\tiny \ 1.73e-8} & \hspace{-1mm} \hbox{\tiny 1.34e-5} & \hspace{-1mm} \hbox{\tiny 0} & \hspace{-1mm} \hbox{\tiny 2.16e-2} & \hspace{-1mm} \hbox{\tiny 20.3} \\ 
 \hspace{-2mm} \hbox{\tiny 0-30} & \hspace{-3mm} \hbox{\tiny 10-20} & \hspace{-3mm} \hbox{\tiny \ acute} & \hspace{-3mm} \hbox{\tiny EC} & \hspace{-1mm} \hbox{\tiny \ 1.25e-10} & \hspace{-1mm} \hbox{\tiny 3.18e-7} & \hspace{-1mm} \hbox{\tiny 0} & \hspace{-1mm} \hbox{\tiny 9.82e-4} & \hspace{-1mm} \hbox{\tiny 34.1} \\ 
 \hspace{-2mm} \hbox{\tiny 0-30} & \hspace{-3mm} \hbox{\tiny 10-20} & \hspace{-3mm} \hbox{\tiny \ obtuse} & \hspace{-3mm} \hbox{\tiny LT} & \hspace{-1mm} \hbox{\tiny \ 6.44e-8} & \hspace{-1mm} \hbox{\tiny 3.56e-5} & \hspace{-1mm} \hbox{\tiny 0} & \hspace{-1mm} \hbox{\tiny 3.58e-2} & \hspace{-1mm} \hbox{\tiny 20.1} \\ 
 \hspace{-2mm} \hbox{\tiny 0-30} & \hspace{-3mm} \hbox{\tiny 10-20} & \hspace{-3mm} \hbox{\tiny \ obtuse} & \hspace{-3mm} \hbox{\tiny EC} & \hspace{-1mm} \hbox{\tiny \ 2.25e-9} & \hspace{-1mm} \hbox{\tiny 4.92e-6} & \hspace{-1mm} \hbox{\tiny 0} & \hspace{-1mm} \hbox{\tiny 1.36e-2} & \hspace{-1mm} \hbox{\tiny 34.1} \\ 
 \hspace{-2mm} \hbox{\tiny 0-30} & \hspace{-3mm} \hbox{\tiny 100-200} & \hspace{-3mm} \hbox{\tiny \ acute} & \hspace{-3mm} \hbox{\tiny LT} & \hspace{-1mm} \hbox{\tiny \ 9.38e-9} & \hspace{-1mm} \hbox{\tiny 4.97e-6} & \hspace{-1mm} \hbox{\tiny 1.75e-15} & \hspace{-1mm} \hbox{\tiny 1.24e-2} & \hspace{-1mm} \hbox{\tiny 18.9} \\ 
 \hspace{-2mm} \hbox{\tiny 0-30} & \hspace{-3mm} \hbox{\tiny 100-200} & \hspace{-3mm} \hbox{\tiny \ acute} & \hspace{-3mm} \hbox{\tiny EC} & \hspace{-1mm} \hbox{\tiny \ 4.20e-11} & \hspace{-1mm} \hbox{\tiny 9.02e-8} & \hspace{-1mm} \hbox{\tiny 0} & \hspace{-1mm} \hbox{\tiny 2.76e-4} & \hspace{-1mm} \hbox{\tiny 31.7} \\ 
 \hspace{-2mm} \hbox{\tiny 0-30} & \hspace{-3mm} \hbox{\tiny 100-200} & \hspace{-3mm} \hbox{\tiny \ obtuse} & \hspace{-3mm} \hbox{\tiny LT} & \hspace{-1mm} \hbox{\tiny \ 7.65e-9} & \hspace{-1mm} \hbox{\tiny 3.40e-6} & \hspace{-1mm} \hbox{\tiny 6.41e-16} & \hspace{-1mm} \hbox{\tiny 6.88e-3} & \hspace{-1mm} \hbox{\tiny 19.3} \\ 
 \hspace{-2mm} \hbox{\tiny 0-30} & \hspace{-3mm} \hbox{\tiny 100-200} & \hspace{-3mm} \hbox{\tiny \ obtuse} & \hspace{-3mm} \hbox{\tiny EC} & \hspace{-1mm} \hbox{\tiny \ 1.78e-11} & \hspace{-1mm} \hbox{\tiny 3.13e-8} & \hspace{-1mm} \hbox{\tiny 1.30e-17} & \hspace{-1mm} \hbox{\tiny 8.26e-5} & \hspace{-1mm} \hbox{\tiny 33.1} \\ 
 
 \hspace{-2mm} \hbox{\tiny 30-60} & \hspace{-3mm} \hbox{\tiny 10-20} & \hspace{-3mm} \hbox{\tiny \ acute} & \hspace{-3mm} \hbox{\tiny LT} & \hspace{-1mm} \hbox{\tiny \ 4.10e-10} & \hspace{-1mm} \hbox{\tiny 3.48e-7} & \hspace{-1mm} \hbox{\tiny 0} & \hspace{-1mm} \hbox{\tiny 6.49e-4} & \hspace{-1mm} \hbox{\tiny 18.2} \\ 
 \hspace{-2mm} \hbox{\tiny 30-60} & \hspace{-3mm} \hbox{\tiny 10-20} & \hspace{-3mm} \hbox{\tiny \ acute} & \hspace{-3mm} \hbox{\tiny EC} & \hspace{-1mm} \hbox{\tiny \ 3.16e-10} & \hspace{-1mm} \hbox{\tiny 9.46e-7} & \hspace{-1mm} \hbox{\tiny 0} & \hspace{-1mm} \hbox{\tiny 2.99e--3} & \hspace{-1mm} \hbox{\tiny 30.6} \\ 
 \hspace{-2mm} \hbox{\tiny 30-60} & \hspace{-3mm} \hbox{\tiny 10-20} & \hspace{-3mm} \hbox{\tiny \ obtuse} & \hspace{-3mm} \hbox{\tiny LT} & \hspace{-1mm} \hbox{\tiny \ 3.04e-9} & \hspace{-1mm} \hbox{\tiny 4.53e-6} & \hspace{-1mm} \hbox{\tiny 0} & \hspace{-1mm} \hbox{\tiny 1.18e-2} & \hspace{-1mm} \hbox{\tiny 18.7} \\ 
 \hspace{-2mm} \hbox{\tiny 30-60} & \hspace{-3mm} \hbox{\tiny 10-20} & \hspace{-3mm} \hbox{\tiny \ obtuse} & \hspace{-3mm} \hbox{\tiny EC} & \hspace{-1mm} \hbox{\tiny \ 6.15e-10} & \hspace{-1mm} \hbox{\tiny 1.40e-6} & \hspace{-1mm} \hbox{\tiny 0} & \hspace{-1mm} \hbox{\tiny 3.95e-3} & \hspace{-1mm} \hbox{\tiny 30.7} \\ 
 \hspace{-2mm} \hbox{\tiny 30-60} & \hspace{-3mm} \hbox{\tiny 100-200} & \hspace{-3mm} \hbox{\tiny \ acute} & \hspace{-3mm} \hbox{\tiny LT} & \hspace{-1mm} \hbox{\tiny \ 6.10e-9} & \hspace{-1mm} \hbox{\tiny 4.69e-6} & \hspace{-1mm} \hbox{\tiny 8.28e-16} & \hspace{-1mm} \hbox{\tiny 9.63e-3} & \hspace{-1mm} \hbox{\tiny 18.2} \\ 
 \hspace{-2mm} \hbox{\tiny 30-60} & \hspace{-3mm} \hbox{\tiny 100-200} & \hspace{-3mm} \hbox{\tiny \ acute} & \hspace{-3mm} \hbox{\tiny EC} & \hspace{-1mm} \hbox{\tiny \ 5.51e-11} & \hspace{-1mm} \hbox{\tiny 8.27e-8} & \hspace{-1mm} \hbox{\tiny 0} & \hspace{-1mm} \hbox{\tiny 2.13e-4} & \hspace{-1mm} \hbox{\tiny 30.7} \\ 
 \hspace{-2mm} \hbox{\tiny 30-60} & \hspace{-3mm} \hbox{\tiny 100-200} & \hspace{-3mm} \hbox{\tiny \ obtuse} & \hspace{-3mm} \hbox{\tiny LT} & \hspace{-1mm} \hbox{\tiny \ 4.55e-9} & \hspace{-1mm} \hbox{\tiny 5.10e-6} & \hspace{-1mm} \hbox{\tiny 1.07e-15} & \hspace{-1mm} \hbox{\tiny 1.33e-2} & \hspace{-1mm} \hbox{\tiny 18.8} \\ 
 \hspace{-2mm} \hbox{\tiny 30-60} & \hspace{-3mm} \hbox{\tiny 100-200} & \hspace{-3mm} \hbox{\tiny \ obtuse} & \hspace{-3mm} \hbox{\tiny EC} & \hspace{-1mm} \hbox{\tiny \ 8.36e-11} & \hspace{-1mm} \hbox{\tiny 2.27e-7} & \hspace{-1mm} \hbox{\tiny 0} & \hspace{-1mm} \hbox{\tiny 7.18e-4} & \hspace{-1mm} \hbox{\tiny 30.7} \\ 
\end{array}$$

 {\tiny (``0" means less that 1e-17 here.)}
 
\ \\ 

Figures 3 - 6 and Table 1 show the results from these experiments. The figures highlight how the calculated errors were statistically distributed. For instance, Figure 4 shows that when the attack angle was restricted to be between 0 and 30 degrees, and the lift amount was restricted to be between 100 and 200, and when the acute triangle was used, the Lambda Twist method had about a quarter of its errors in the $10^{-12}$ to $10^{-13}$ range, while the elliptic curve method had nearly half of its errors in this range. 

Table 1 presents more details concerning these tests. It shows, for each experiment, the average error and its standard deviation, the minimum error, the maximum error, and the execution time for the combined ten million trials. Remember that the errors being discussed here are relative errors, based on taking absolute distance errors and dividing by distances from the camera to the control points. 

Obviously, the Lambda Twist method is much faster. Nevertheless, the elliptic curve method tends to be more accurate as long as the attack angle is not too great. The error differences between the elliptic curve (EC) method and the Lambda Twist (LT) method are often relatively large. Their absolute differences may be fairly small, but again, these errors are themselves relative errors. Still, these differences might be too small to be of practical use, given the accuracies of the devices currently used for physical measurements in situations where P3P might reasonably be employed.  Once again, the reader is invited to obtain the previously mentioned C program to check the results, conduct further experiments and scrutinize the algorithm implementations.

\section{\uppercase{Conclusion}}
\label{sec:Conclusion}

The classical Perspective Three-Point (P3P) problem has been shown to have an intriguing connection with a certain problem involving sliding an arc around on the unit sphere. As a consequence of this, it is  equivalent to the problem of finding the intersection of a quartic curve and a line, in the projective plane. An algorithm for exploiting this to solve the P3P problem has been presented and investigated. In practice, it yields highly accurate results, but is considerably slower than a certain state-of-the-art P3P solver. The connection between P3P and quartic curves should be further investigated. This would likely produce useful insights into the P3P problem, such as understanding better how its solutions are related, and possibly leading to a better P3P solver. 

A rather general family of genus-one quartic planar curves was introduced, along with a simple birational transformation to convert these to Jacobi quartic curves. The formulas involved are strikingly simple and symmetric in the coefficients of the polynomial for the original curve. Edwards and twisted Edwards curves belong to this family, as do the curves associated with the P3P problem. This family of curves should be further investigated. In particular, the present work completely ignored the Abelian group structures of these curves, which might well be a fruitful line of inquiry.

 \section*{Acknowledgement}

The author wishes to acknowledge the contributions of Mr. Pawel Barnas. He is responsible for capably designing and implementing Python testing software for the P3P-solving code developed by the author. He also ran tests and collected the data.

\renewcommand{\baselinestretch}{0.98}

\bibliographystyle{apalike}

\renewcommand{\baselinestretch}{1}

\end{document}